\documentclass{article} 
\usepackage{iclr2027_conference,times}

\usepackage{amsmath,amsfonts,bm}

\def\eqref#1{equation~\ref{#1}}

\def\1{\bm{1}}

\DeclareMathAlphabet{\mathsfit}{\encodingdefault}{\sfdefault}{m}{sl}
\SetMathAlphabet{\mathsfit}{bold}{\encodingdefault}{\sfdefault}{bx}{n}

\usepackage{hyperref}
\usepackage{url}

\usepackage{graphicx}
\usepackage{algorithm}
\usepackage{algorithmic}
\usepackage{booktabs}
\usepackage{multirow}
\usepackage{colortbl}
\usepackage{amsmath}
\usepackage{amssymb}
\usepackage{xcolor}
\usepackage{pgfplots}
\usepackage{wrapfig}
\pgfplotsset{compat=1.17}
\usetikzlibrary{positioning, arrows.meta}
\usepackage{tcolorbox, pifont}

\title{DREAM: \textbf{D}ynamic \textbf{Re}solution \textbf{A}ssignment for \textbf{M}ultimodal Multi-Agent Debate}

\iclrfinalcopy

\author{Khanh-Binh~Nguyen, Van~Dai~Do, Tien~Anh~Nguyen, Svetha~Venkatesh \& Hung~Le\\
Deakin University\\
Applied Artificial Intelligence Initiative, Geelong, Australia \\
\texttt{\{kbi,v.do,tien.nguyen,svetha.venkatesh,thai.le\}@deakin.edu.au}
}

\begin{document}

\maketitle

\begin{abstract}
Multi-agent debate (MAD) has emerged as an effective paradigm to improve the reasoning capabilities of large language models (LLMs) and is increasingly being extended to multimodal settings. However, existing multimodal MAD frameworks typically expose agents to the same fixed visual input, ignoring substantial variation in the visual scale needed across samples and agents. In addition, these frameworks frequently suffer from \textit{groupthink}—a phenomenon where agents prematurely abandon correct deductions to conform with confident but hallucinated peer responses. To address these bottlenecks, we introduce \textbf{DREAM} (\textbf{D}ynamic \textbf{Re}solution \textbf{A}ssignment For \textbf{M}ultimodal Multi-Agent Debate), which operates via two core components: (1) Dynamic Resolution Assignment, a zero-shot ``probe round'' where agents test multiple resolutions, quantify uncertainty using Average Normalized Log-Likelihood (ANLL), and use an adaptive threshold to assign each agent to its empirically optimal resolution; (2) Uncertainty-Guided Rollback Aggregation counters groupthink by tracking each agent’s uncertainty over rounds and restoring early low-uncertainty answers overridden by group pressure. On six multimodal datasets, DREAM improves the accuracy–token trade-off over multi-agent debate baselines by 1.5–3.2\% accuracy without dataset-specific tuning.
\end{abstract}

\section{Introduction}

\begin{wrapfigure}[23]{r}{0.5\textwidth}
    \vspace{-\intextsep}
    \hspace*{-.75\columnsep}
    \includegraphics[width=\linewidth]{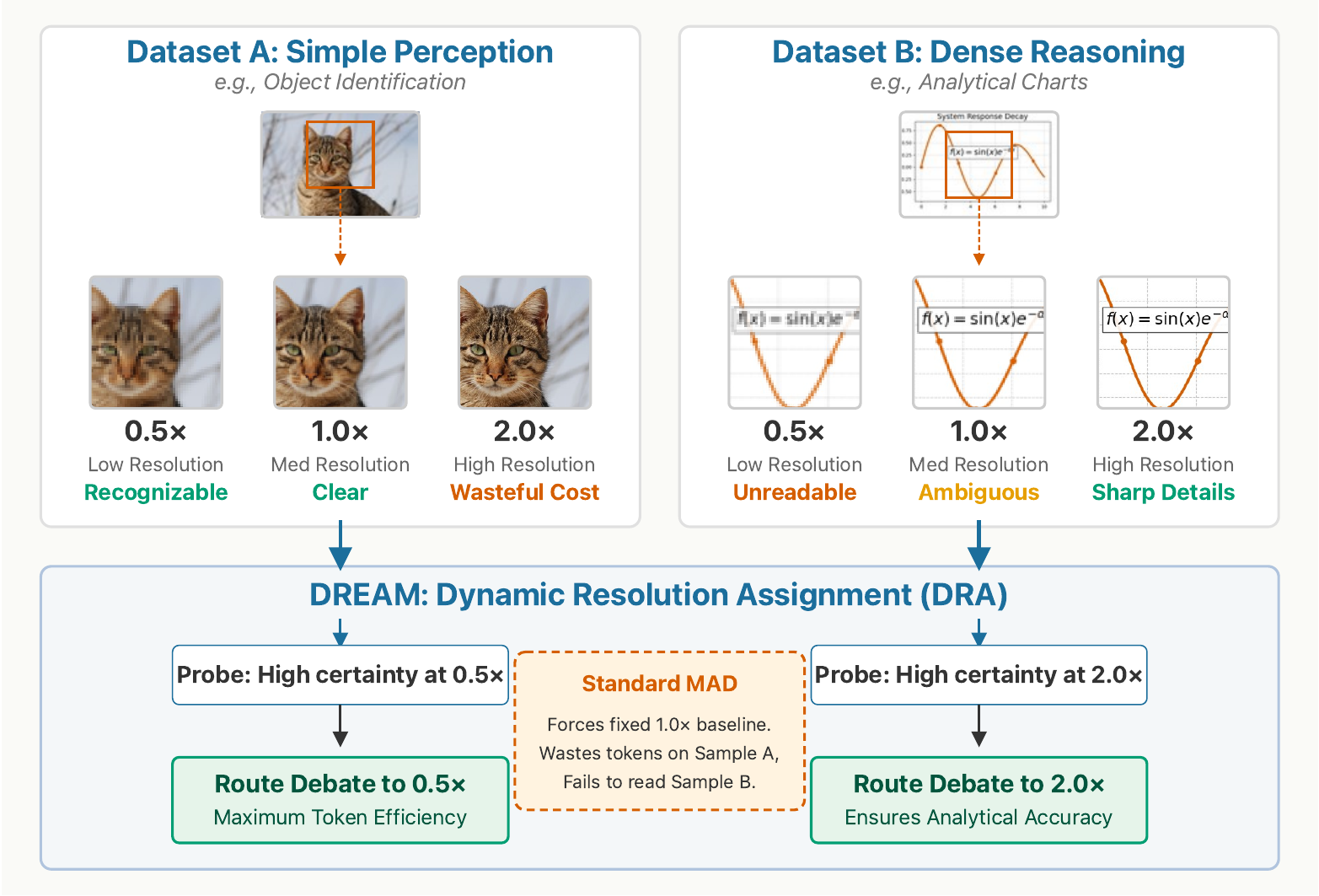}
    \caption{Visual tasks demand different resolutions: simple perception tasks (left) only need $0.5\times$, while $2.0\times$ introducing long-context noise that actively degrades accuracy. Conversely, dense charts (right) require $2.0\times$ to preserve critical fine-grained details, thus $0.5\times$ irreversibly discarding them. Standard MAD uses a fixed resolution, degrading on simple samples and failing on complex ones. DREAM solves this by dynamically assigning each sample to its empirically optimal scale.}
    \label{fig:motivation}
\end{wrapfigure}



Reasoning in large language models (LLMs) has advanced rapidly, with significant progress driven by the orchestration of multiple model instances. Multi-agent debate (MAD) \citep{debate1, debate2} has emerged as a widely adopted paradigm where agents iteratively exchange answers and critiques, revising their responses under the assumption that logical errors are more easily detected via peer review. Recently, MAD has been extended to multimodal reasoning, standardizing visual collaboration in benchmarks like M3MAD-Bench \citep{li2026m3mad}. However, while debate effectively refines textual logic, directly porting it to vision-language tasks exposes severe structural bottlenecks, specifically regarding visual scale and consensus integrity.

The first major bottleneck in existing frameworks is the static handling of visual resolution. Although modern Multimodal LLMs (MLLMs) \citep{openai2023gpt4v, geminiteam2024gemini, bai2023qwenvl} natively support dynamic resolution encoding, current multi-agent systems force a rigid, one-size-fits-all visual scale. Simply maximizing resolution is inadequate: excessively high resolutions on simple perception tasks actively degrade performance by injecting long-context noise and spurious features \citep{wang2024qwen2vl, dong2024internlmxcomposer2_4khd}, while low resolutions obscure critical text in dense documents (Figure~\ref{fig:motivation}). Because the optimal visual scale is highly sample-dependent \citep{niu2025nativevisual, li2025resbench}, static assignments leave agents fundamentally ungrounded.
Crucially, this loss of visual grounding directly deepens a second bottleneck: \textit{sycophantic groupthink}. Standard MAD inherently suffers from herding \citep{wynn2025talk, sharma2023sycophancy, xiong2023examining}, and multimodality amplifies this vulnerability. When suboptimal resolution deprives agents of the ability to independently verify visual claims, a single hallucinated assertion can easily trigger a cascade of blind agreement. Under this peer pressure, valid early deductions are displaced, which is a phenomenon measurable as certainty erosion, and the final consensus becomes dominated by sycophancy rather than reasoned deduction.

To jointly resolve these structural issues, this work introduces \textbf{DREAM} (\textbf{D}ynamic \textbf{Re}solution \textbf{A}ssignment for \textbf{M}ultimodal Multi-Agent Debate), a plug-and-play wrapper featuring two complementary mechanisms: 
(1)~\textit{Dynamic Resolution Assignment (DRA)}: An exploratory ``Round~0'' probe dynamically determines the optimal visual scale for each sample using Average Normalized Log-Likelihood (ANLL) as a zero-shot uncertainty signal. Based on initial consensus, DRA assigns agents to either a locked single scale to minimize noise, or a mixed-resolution payload for maximal visual coverage. 
(2)~\textit{Uncertainty-Guided Rollback Aggregation (URA)}: A targeted defense against late-stage groupthink that monitors ANLL trajectories throughout the debate. If an agent's uncertainty sharply rises by the final round, URA rescues their most confident prior response. To counteract judge recency bias, rescued answers are strategically prepended to the final prompt, ensuring early, visually grounded deductions are weighed equally against the prevailing groupthink.

Our main contributions are summarized as follows:
\begin{itemize}
    \item We identify and formally characterize two structural bottlenecks universal to all existing multimodal MAD frameworks: static visual resolution assignment and late-stage sycophantic groupthink, and demonstrate how these two challenges are causally intertwined.
    
    \item We propose DREAM, to our knowledge the first multimodal MAD framework to jointly resolve these bottlenecks. DRA introduces a zero-shot uncertainty-driven probing phase that dynamically routes each sample to its empirically optimal resolution strategy, while URA provides a principled rollback mechanism that prevents correct early deductions from being displaced by group pressure.
    
    \item We conduct extensive experiments across six multimodal benchmarks and four open-weight MLLMs, demonstrating that DREAM consistently improves the accuracy-token trade-off as a plug-and-play wrapper over existing debate baselines and other multimodal agentic approaches, without requiring any dataset-specific tuning.
\end{itemize}
\section{Related Work}

\subsection{Multi-Agent Debate for LLMs}
\citet{debate1} introduced multi-agent debate (MAD) to enhance factual accuracy by allowing multiple LLM instances to critique and refine each other's responses. Subsequent research has diversified these structures \citep{debate2, liu2025breaking}, though some explicitly question whether debate consistently outperforms advanced single-model strategies \citep{wang2024rethinking, smit2024should}. Across all configurations, a persistent vulnerability remains groupthink, where agents converge on an answer regardless of correctness due to herding and sycophancy \citep{sharma2023sycophancy, wynn2025talk}. While recent approaches attempt to mitigate this using voting mechanisms, verbal exchanges, or confidence-weighted updates \citep{benedikt2025voting, lin2025confmad, tian2026mad_mm}, they still treat uncertainty strictly as an intra-debate filter. Crucially, none of these methods track how an individual agent's uncertainty evolves across rounds, nor do they leverage it to dynamically assign the optimal visual inputs prior to the debate.

MAD has also been extended to multimodal settings, including deception detection \citep{FangDEBATEMODELS} and visual misinformation \citep{mad_sherlock}, while \citet{HuangBeCollaboration} separate perception from reasoning via a dedicated perceiver agent. M3MAD-Bench \citep{li2026m3mad} further standardizes multimodal debate and highlights substantial token inefficiencies. Critically, all existing approaches operate under static initial conditions, assigning identical visual inputs to every agent and leaving per-sample resolution assignment entirely unaddressed. To our knowledge, DREAM is the first multimodal MAD framework to treat resolution as a per-agent, per-sample decision and to jointly counter groupthink by tracking each agent's uncertainty trajectory rather than relying on fixed aggregation or prompting strategies.

\subsection{Uncertainty and Confidence in Multi-Agent Systems}
Single-model uncertainty estimation spans token-level self-knowledge, semantic uncertainty, and hallucination detection \citep{kuhn2023semantic, xiong2024llmuncertainty}. In multi-agent settings, uncertainty metrics are primarily used to filter message propagation between agents \citep{lin2025confmad, tian2026mad_mm}, although such filtering can inadvertently restrict output diversity \citep{nguyen2026creativemad}. However, a critical connection between uncertainty and groupthink remains unexploited: when an agent is pressured into abandoning a visually grounded answer, its confidence in the coerced response drops, presenting as a measurable rise in uncertainty across debate rounds. In other words, groupthink leaves a detectable uncertainty trace. DREAM leverages this phenomenon in two distinct ways: first, as a zero-shot pre-debate signal to route agents to their optimal visual resolution; and second, as a per-round trajectory tracker that identifies compromised agents, rescuing their most confident earlier deduction when late-stage uncertainty exceeds a calibrated threshold.
\section{Methodology}
\label{sec:method}

\begin{figure}[t]
    \centering
    \includegraphics[width=0.8\linewidth]{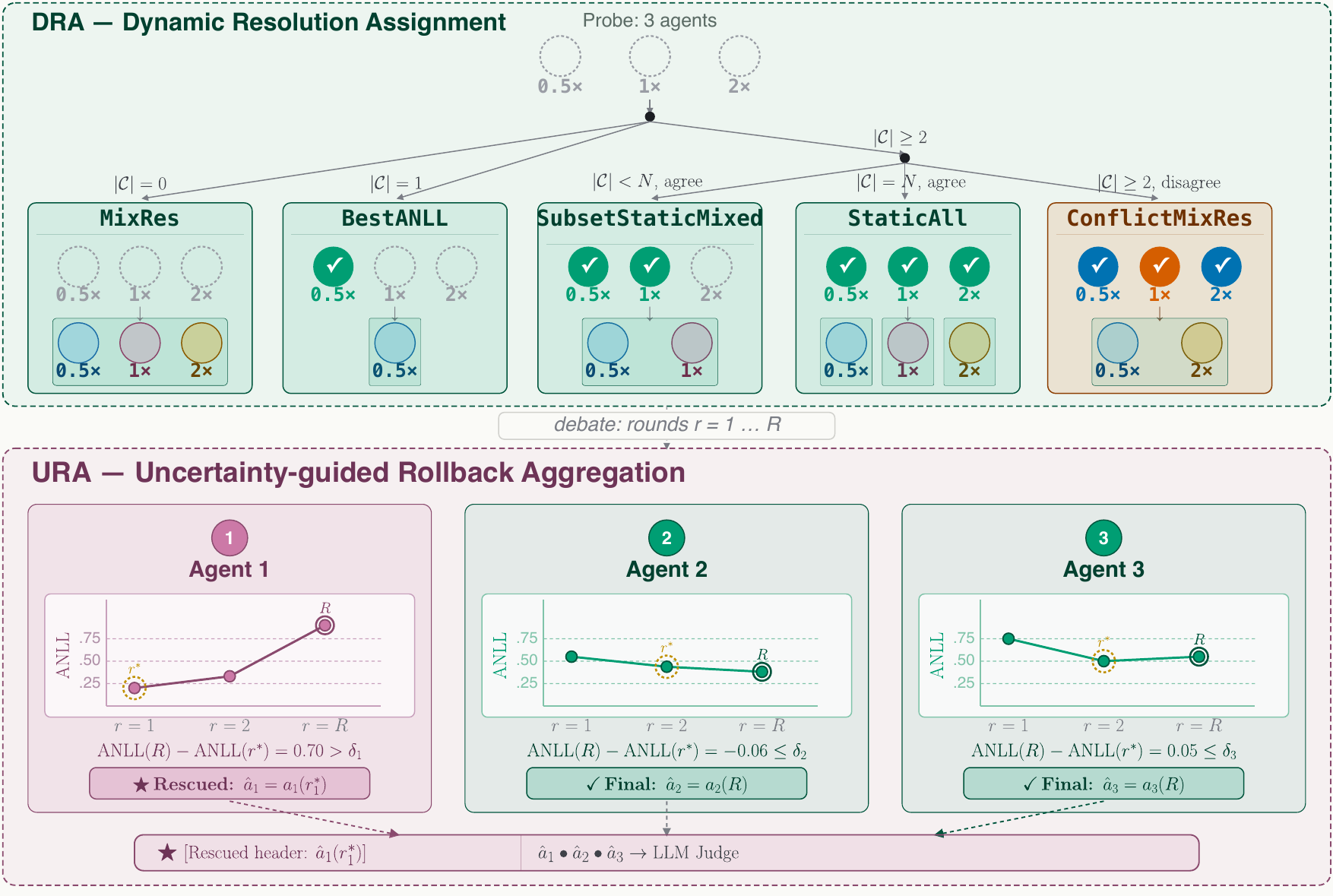}
    \caption{\textbf{The DREAM framework pipeline.} \textbf{Top:} DRA's probe-round decision tree, selecting a DRA strategy based first on confident-set size $|\mathcal{C}|$ and then, when $|\mathcal{C}| \geq 2$, on answer agreement within $\mathcal{C}$; the disagreement strategy (\textsc{ConflictMixRes}) is highlighted in orange. \textbf{Bottom:} URA's per-agent ANLL trajectory over debate rounds $r=1,2,\dots,R$, comparing each agent's final round to its earlier peak $r^*$ (gold marker) against the rescue threshold $\delta_i$. Agent~1 (pink) exceeds $\delta_1$ and is rescued; Agents~2--3 (green) keep their final answers. Green marks the default path throughout.}
    \label{fig:pipeline}
\end{figure}

We introduce DREAM, a multi-agent multimodal debate framework comprising two key innovations: Dynamic Resolution Assignment (DRA) and Uncertainty-Guided Rollback Aggregation (URA). DRA conducts a preliminary round to identify the most suitable resolution for each agent based on the current sample. Meanwhile, URA monitors each agent's responses throughout the debate, allowing it to recover the low-uncertainty answer that might have been overlooked due to groupthink.
Figure~\ref{fig:pipeline} details the branching logic of the Route step and how its output feeds the unchanged debate phase and the rollback aggregation step. The full four-phase procedure (Probe $\rightarrow$ Route $\rightarrow$ Debate $\rightarrow$ Aggregate \& Judge) is given as Algorithm~\ref{alg:DREAM}.

\subsection{Overview}

Given a visual question-answering sample $(I, q)$, a standard MLLM $\mathcal{M}$ produces an answer $a = \mathcal{M}(I_s, q)$, where $I_s$ denotes image $I$ rendered at a resolution scale $s$ drawn from a discrete candidate set $\mathcal{S} = \{s_1, s_2, \ldots, s_K\}$. With $\mathcal{M}$, $q$, and $I$ held constant, both the model's answer and its output uncertainty vary with $s$, and the direction and magnitude of this change are sample-dependent. Naively assigning the highest resolution to all agents is suboptimal because different tasks and image types benefit from different scales; likewise, feeding \emph{all} resolutions to \emph{every} agent by default inflates the visual token count and can introduce noise-induced hallucinations (we therefore reserve multi-scale inputs for cases where probing indicates that no single scale is reliably confident).

In a multi-agent debate, $N$ agents $\{\mathcal{M}_i\}_{i=1}^N$ exchange candidate answers over $R$ rounds. At round $r$, agent $i$ receives the image at its assigned resolution $s_i^{(r)} \in \mathcal{S}$, its own prior response, and its peers' responses, and produces an updated answer. In standard MAD, $s_i^{(r)}$ is the same fixed value for all $i$ and all $r$. DREAM relaxes this: DRA selects a per-agent, per-sample resolution $s_i^*$ in a preliminary probe round. More importantly, URA tracks each agent's certainty trajectory $\{\mathrm{ANLL}_i^{(r)}\}_{r=1}^R$ across rounds to identify and rescue certainty-diminished answers at aggregation time.

The output certainty of each agent is measured using the ANN. ANLL is preferred because it is straightforward, length-normalized, and inexpensive to compute from the model's token log-probabilities. This makes it easy to track throughout each debate round without requiring additional forward passes or auxiliary models. Although there are many alternative uncertainty metrics available, such as those based on ensembles, self-consistency, or calibrated sampling, we choose ANLL as a lightweight and broadly applicable measure of generation confidence.

\begin{equation}
    \mathrm{ANLL}_i^{(r)} = -\frac{1}{|y_i^{(r)}|} \sum_{t=1}^{|y_i^{(r)}|} \log p_\theta\!\left(y_{i,t}^{(r)} \,\middle|\, y_{i,<t}^{(r)},\, x_i^{(r)}\right)
    \label{eq:anll}
\end{equation}
where $x_i^{(r)}$ is the agent's input at round $r$ and $y_i^{(r)}$ is its generated response. A lower ANLL indicates that the model assigned high probability to every token it generated.

\subsection{Dynamic Resolution Assignment}

\noindent\textbf{Probe round.}
Before any debate exchange occurs, DREAM inserts a \emph{Round~0} whose history is discarded and never shared with peers. Each agent $i$ is assigned a distinct resolution $s_i^{(0)} \in \mathcal{S}$ from the candidate set in round-robin order, so the $N$ agents collectively span the resolution range. Every agent answers the question independently at its assigned resolution using greedy decoding (temperature $= 0$), and we record both its extracted answer $\hat{a}_i^{(0)}$ and its $\mathrm{ANLL}_i^{(0)}$.

\noindent\textbf{Adaptive confidence threshold.}
Given the probe ANLL vector $\{\mathrm{ANLL}_i^{(0)}\}_{i=1}^N$, DREAM computes an effective confidence threshold $\tau_{\mathrm{eff}}$ that adapts to the sample's difficulty. We examine two thresholding schemes. Under \emph{static} thresholding, $\tau_{\mathrm{eff}} = \tau$ is a fixed scalar (the default is set to 0.2). Under \emph{adaptive} thresholding, the threshold is set to one standard deviation below the mean of the probe-round ANLL scores:
\begin{equation}
    \mu_0 = \frac{1}{N}\sum_{i=1}^{N}\mathrm{ANLL}_i^{(0)},
    \quad
    \sigma_0 = \sqrt{\frac{1}{N}\sum_{i=1}^{N}\!\left(\mathrm{ANLL}_i^{(0)} - \mu_0\right)^2},
    \quad
    \tau_{\mathrm{eff}} = \mu_0 - \sigma_0
    \label{eq:threshold}
\end{equation}
Agents whose ANLL falls at least one standard deviation below the sample mean are classified as genuinely confident; those above $\tau_{\mathrm{eff}}$ are not. This formulation is self-calibrating: on easy samples where all ANLLs are low and tightly clustered, $\tau_{\mathrm{eff}}$ is also low and the bar is appropriately strict; on hard samples with high, spread-out ANLLs, $\tau_{\mathrm{eff}}$ rises accordingly. When fewer than two valid probe scores are available, DREAM falls back to the static threshold $\tau$. Agent $i$ is classified as \emph{confident} if $\mathrm{ANLL}_i^{(0)} \leq \tau_{\mathrm{eff}}$.

\noindent\textbf{Uncertainty-Guided Resolution Assignment.}
After the probe round, each agent $i$ holds an answer $\hat{a}_i^{(0)}$ and an uncertainty score $\mathrm{ANLL}_i^{(0)}$. Agents whose score falls below the adaptive threshold $\tau_{\mathrm{eff}}$ form the \textit{confident set} $\mathcal{C} = \{i : \mathrm{ANLL}_i^{(0)} \leq \tau_{\mathrm{eff}}\}$, and their probe resolutions are collected as $\mathcal{S}^*_\mathcal{C} = \{s_i^{(0)} : i \in \mathcal{C}\}$. The composition and consensus of $\mathcal{C}$ then determine which resolution payload each agent receives for all subsequent debate rounds:
\begin{enumerate}
    \item \textbf{No confident agent ($|\mathcal{C}| = 0$).} No resolution produced a reliable signal. All agents receive the full multi-scale payload $\mathcal{S}$ (\textsc{MixRes}).

    \item \textbf{Exactly one confident agent ($|\mathcal{C}| = 1$).} A single resolution has proven reliable. All agents are locked to that agent's probe resolution (\textsc{BestANLL}).

    \item \textbf{All agents confident and in agreement ($|\mathcal{C}| = N$ and $\hat{a}_i^{(0)} = \hat{a}_j^{(0)} \; \forall i, j \in \mathcal{C}$).} Every agent both converged and agreed. Each agent retains its own probe resolution, preserving the diversity across scales established during probing (\textsc{StaticAll}).

    \item \textbf{A strict confident subset, in agreement ($0 < |\mathcal{C}| < N$ and $\hat{a}_i^{(0)} = \hat{a}_j^{(0)} \; \forall i, j \in \mathcal{C}$).} Confident agents have identified reliable resolutions, but unconfident agents have not. Confident agents retain their own probe resolutions, while unconfident agents receive the mixture of $\mathcal{S}^*_\mathcal{C}$, exposing them to the scales that produced reliable signals in their peers. When all confident agents were probed at the same scale, this reduces to a single locked resolution (\textsc{SubsetStaticMixed}~/~\textsc{BestANLL}).

    \item \textbf{Confident agents are divided ($|\mathcal{C}| \ge 2$ and $\exists i, j \in \mathcal{C} \text{ s.t. } \hat{a}_i^{(0)} \neq \hat{a}_j^{(0)}$).} Agreement failed despite low uncertainty, signaling that the query is genuinely ambiguous across scales. A majority vote is taken over the confident agents' answers. If a strict majority exists, the minority resolutions are discarded, and all agents receive the majority-supporting resolutions $\mathcal{S}^\dagger \subseteq \mathcal{S}^*_\mathcal{C}$ as a single scale or mixture. If no strict majority exists, all of $\mathcal{S}^*_\mathcal{C}$ is used as a mixture to maximize visual coverage (\textsc{ConflictMixRes}).
\end{enumerate}

In all cases, the probe history is discarded before the debate begins, ensuring agents enter without anchoring bias from Round~0 responses.

\subsection{Uncertainty-Guided Rollback Aggregation}
\noindent\textbf{Per-agent ANLL trajectory.}
During every debate round $r \in \{1, \ldots, R\}$, DREAM records each agent's ANLL score alongside its response, forming a trajectory $\{\mathrm{ANLL}_i^{(r)}\}_{r=1}^R$. This trajectory captures how certainty evolves under peer pressure: an agent that begins confident but is persuaded by an incorrect argument will typically exhibit a rising ANLL in later rounds, which is a signal of \emph{certainty erosion}.

\noindent\textbf{Debate-calibrated rescue threshold.}
For each agent $i$, we compute the standard deviation of its ANLL trajectory across the $R$ debate rounds:
\begin{equation}
    \sigma_i^{\mathrm{debate}} = \sqrt{\frac{1}{R}\sum_{r=1}^{R}\!\left(\mathrm{ANLL}_i^{(r)} - \bar{\mu}_i\right)^2}, \quad \bar{\mu}_i = \frac{1}{R}\sum_{r=1}^{R}\mathrm{ANLL}_i^{(r)}
    \label{eq:sigma_debate}
\end{equation}
The per-agent rescue threshold is then $\delta_i = \alpha \cdot \sigma_i^{\mathrm{debate}}$, where $\alpha$ is a scalar hyperparameter (default $0.5$). A larger $\sigma_i^{\mathrm{debate}}$ means agent $i$'s confidence fluctuated widely during debate, so $\delta_i$ is raised to avoid triggering rescue on natural variation. A small $\sigma_i^{\mathrm{debate}}$ means the agent was consistently certain throughout, so $\delta_i$ is tight, and any late-round erosion is readily detected.

\noindent\textbf{Answer rescue.}
At aggregation time, let $r^*_i = \arg\min_{r < R} \mathrm{ANLL}_i^{(r)}$ be agent $i$'s most confident round strictly before the final. Rescue triggers for agent $i$ if:
\begin{equation}
    \mathrm{ANLL}_i^{(R)} - \mathrm{ANLL}_i^{(r^*_i)} > \delta_i
    \label{eq:rescue}
\end{equation}
When rescue triggers, agent $i$'s contribution is replaced by its answer from round $r^*_i$ rather than round $R$. By construction, rescue can only point to an earlier round.

\noindent\textbf{Anti-recency bias.}
The judge receives exactly one answer per agent: the rescued answer $a_i^{(r^*_i)}$ if rescue was triggered, or the final-round answer $a_i^{(R)}$ otherwise. Since LLM judges tend to overweight answers that appear later in the prompt due to position bias, DREAM prepends a structured header for each rescued answer at the top of the judge's prompt:

\begin{quote}
\texttt{[Agent $i$ Rescued Answer (Round $r^*_i$): ``\textit{answer}'']}
\end{quote}

The judge is then instructed to weigh rescued answers equally with all others, ensuring that an earlier, lower-uncertainty answer is not discounted simply because it originates from a prior round.

\section{Experimental Setup}
\label{sec:setup}

\textbf{Datasets.}
The DREAM framework is evaluated on the multimodal evaluation subset of M3MAD-Bench \citep{li2026m3mad}: six benchmarks spanning knowledge (MME), mathematics (MathVista, MathVision), medicine (PathVQA), and reasoning (MME-Reasoning, VisualPuzzles). Appendix~\ref{sec:appendix_datasets} provides full dataset descriptions and the experimental setup.

\textbf{Models.}
To assess the framework across diverse architectures and model sizes, the experiments employ four primary multimodal large language models (MLLMs): \textbf{Gemma-3-4B}, \textbf{Qwen2.5-VL-7B}, \textbf{InternVL3-8B}, and \textbf{GPT-4o-mini}. These models form the core analysis presented in Table~\ref{tab:main_multimodal}.
More evaluated models are in Appendix Table~\ref{tab:appendix_multimodal}.

\textbf{Baselines.}
Comparisons are made against standard inference methods and existing multi-agent frameworks. Single-agent baselines include \textbf{IO} (direct answer), \textbf{CoT} \citep{wei2022chain}, and \textbf{SC} \citep{sc} at native resolution. Multi-agent debate baselines include \textbf{LLM Debate} \citep{debate1}, \textbf{Div-MAD} \citep{debate2}, \textbf{DMAD} \citep{liu2025breaking}, and \textbf{Debate with Images (DwI)} \citep{FangDEBATEMODELS}. Finally, to isolate dynamic resolution assignment from the underlying debate logic, the textual debate baselines are evaluated under three distinct \textbf{Resolution Wrappers}: bare (fixed resolution), \textbf{Perceiver} \citep{HuangBeCollaboration}, and the proposed \textbf{DREAM} framework. 

\textbf{Implementation Details.}
All experiments utilize greedy decoding (temperature $= 0$) for the initial probe round to ensure deterministic, reliable uncertainty measurement, and temperature $= 0.5$ for subsequent debate rounds to encourage conversational diversity. The rescue scaling hyperparameter $\alpha = 0.5$ is fixed across all datasets; the per-agent rescue threshold is defined as $\delta_i = \alpha \cdot \sigma_i^{\mathrm{debate}}$, where $\sigma_i^{\mathrm{debate}}$ is the standard deviation of agent $i$'s ANLL trajectory across the debate rounds (Eq.~\ref{eq:sigma_debate}). ANLL is computed over the full response sequence, excluding special tokens. To rigorously account for generation variance during the debate rounds, all reported metrics represent the mean accuracy computed across three independent random seeds. Full implementation details are reported in Appendix~\ref{app:implementation}.

\section{Results and Analysis}
\label{sec:results}

\begin{table*}[t]
    \centering
    \caption{Performance on M3MAD-Bench across four distinct model families (results for an additional model are provided in Appendix Table~\ref{tab:appendix_multimodal}). Best results are \textbf{bolded}, second-best \underline{underlined}. $\Delta$ denotes DREAM's average improvement over the LLM Debate baseline. \textit{MVt}: MathVista, \textit{MVs}: MathVision, \textit{PVQA}: PathVQA, \textit{MME-R}: MME-Reasoning, \textit{VP}: VisualPuzzles.}
    \label{tab:main_multimodal}
    \footnotesize
    \setlength{\tabcolsep}{6pt}
    \begin{tabular}{c l c cc c cc c c}
    \toprule
        \multirow{2}{*}{\textbf{Model}} &
        \multirow{2}{*}{\textbf{Method}} &
        \textbf{Know.} &
        \multicolumn{2}{c}{\textbf{Mathematics}} &
        \textbf{Med.} &
        \multicolumn{2}{c}{\textbf{Reasoning}} &
        \multirow{2}{*}{\textbf{Avg.}} &
        \multirow{2}{*}{$\mathbf{\Delta}$} \\
        \cmidrule(lr){3-3} \cmidrule(lr){4-5} \cmidrule(lr){6-6} \cmidrule(lr){7-8}
        & & \textbf{MME} & \textbf{MVt} & \textbf{MVs} & \textbf{PVQA} & \textbf{MME-R} & \textbf{VP} & & \\
    \midrule
    \multirow{9}{*}{\rotatebox[origin=c]{90}{\textcolor{cyan!80!black}{\textbf{Gemma-3-4B}}}}
        & IO & 79.3 & 37.4 & 15.0 & 24.9 & 21.8 & 22.9 & 33.5 & \\
        & CoT~[\citenum{wei2022chain}] & 80.2 & 37.4 & 15.7 & 24.9 & 21.8 & 22.9 & 33.8 & \\
        & SC~[\citenum{sc}] & 80.1 & 37.1 & 15.1 & 24.9 & 20.8 & 22.8 & 33.5 & \\
        & DwI~[\citenum{FangDEBATEMODELS}] & 58.9 & 23.7 & 8.7 & 26.1 & 23.0 & 18.6 & 26.5 & \\
        & LLM Debate~[\citenum{debate1}] & 78.4 & \underline{57.0} & \underline{24.3} & \underline{40.6} & \underline{26.5} & \underline{36.3} & \underline{43.9} & \\
        & Div-MAD~[\citenum{debate2}] & 32.7 & 32.7 & 17.8 & 17.6 & 19.7 & 26.3 & 24.5 & \\
        & DMAD~[\citenum{liu2025breaking}] & 79.8 & 51.6 & 21.6 & 39.5 & 25.4 & 36.1 & 42.3 & \\
        & Perceiver~[\citenum{HuangBeCollaboration}] & \textbf{82.5} & 52.4 & 23.1 & 35.0 & 26.0 & 32.7 & 41.9 & \\
        \cmidrule(l){2-10}
        & \textbf{DREAM} & \underline{80.7} & \textbf{57.3} & \textbf{27.3} & \textbf{41.6} & \textbf{28.2} & \textbf{37.0} & \textbf{45.4} & \textbf{\textcolor{green!50!black}{$\uparrow$1.5}} \\
    \midrule
    \multirow{9}{*}{\rotatebox[origin=c]{90}{\textcolor{orange!80!black}{\textbf{Qwen2.5-VL-7B}}}}
        & IO & 73.4 & 64.8 & 25.2 & \underline{45.4} & 27.8 & 34.4 & 45.2 & \\
        & CoT~[\citenum{wei2022chain}] & 67.8 & 66.6 & 23.0 & 26.0 & 32.0 & 34.8 & 41.7 & \\
        & SC~[\citenum{sc}] & \underline{76.8} & 67.0 & \underline{27.8} & 27.0 & 30.0 & 36.8 & 44.2 & \\
        & DwI~[\citenum{FangDEBATEMODELS}] & 73.2 & 67.3 & 20.8 & 26.9 & 26.8 & 33.6 & 41.4 & \\
        & LLM Debate~[\citenum{debate1}] & 72.3 & \underline{68.3} & 26.3 & 42.2 & \underline{33.2} & \underline{39.5} & 47.0 & \\
        & Div-MAD~[\citenum{debate2}] & 57.8 & 62.2 & 24.2 & 17.1 & 26.4 & 33.2 & 36.8 & \\
        & DMAD~[\citenum{liu2025breaking}] & 74.2 & 68.6 & 27.1 & \textbf{46.2} & 32.3 & 35.3 & \underline{47.3} & \\
        & Perceiver~[\citenum{HuangBeCollaboration}] & 54.1 & 65.4 & 24.4 & 39.1 & 30.4 & 38.9 & 42.1 & \\
        \cmidrule(l){2-10}
        & \textbf{DREAM} & \textbf{81.9} & \textbf{73.3} & \textbf{28.5} & 41.6 & \textbf{34.5} & \textbf{41.6} & \textbf{50.2} & \textbf{\textcolor{green!50!black}{$\uparrow$3.2}} \\
    \midrule
    \multirow{9}{*}{\rotatebox[origin=c]{90}{\textcolor{violet!80!black}{\textbf{InternVL3-8B}}}}
        & IO & 85.0 & 65.8 & 27.0 & \textbf{52.0} & 30.4 & \underline{38.2} & \underline{49.7} & \\
        & CoT~[\citenum{wei2022chain}] & 85.0 & 66.6 & 27.6 & \underline{44.4} & 29.0 & 34.8 & 47.9 & \\
        & SC~[\citenum{sc}] & 86.0 & 65.0 & \underline{29.0} & 32.0 & 28.8 & 37.8 & 46.4 & \\
        & DwI~[\citenum{FangDEBATEMODELS}] & 81.6 & 66.3 & 23.6 & 38.5 & \underline{30.9} & 37.3 & 46.4 & \\
        & LLM Debate~[\citenum{debate1}] & \underline{88.6} & \underline{70.6} & 28.7 & 37.4 & 30.8 & 37.5 & 48.9 & \\
        & Div-MAD~[\citenum{debate2}] & 42.4 & 44.7 & \underline{29.0} & 32.6 & 26.3 & 31.3 & 34.4 & \\
        & DMAD~[\citenum{liu2025breaking}] & 85.2 & 67.2 & \textbf{29.4} & 42.2 & 29.8 & 37.0 & 48.5 & \\
        & Perceiver~[\citenum{HuangBeCollaboration}] & 84.8 & 70.0 & 25.5 & 39.1 & 30.7 & 34.7 & 47.5 & \\
        \cmidrule(l){2-10}
        & \textbf{DREAM} & \textbf{89.6} & \textbf{73.3} & \textbf{29.4} & 42.9 & \textbf{31.6} & \textbf{41.7} & \textbf{51.4} & \textbf{\textcolor{green!50!black}{$\uparrow$2.5}} \\
    \midrule
    \multirow{7}{*}{\rotatebox[origin=c]{90}{\textcolor{green!70!black}{\textbf{GPT-4o-mini}}}}
        & IO & 72.6 & 57.0 & \underline{32.0} & \underline{39.6} & 26.0 & 38.8 & 44.3 & \\
        & CoT~[\citenum{wei2022chain}] & 65.2 & 57.6 & 26.2 & 14.4 & \underline{30.0} & 37.4 & 38.5 & \\
        & SC~[\citenum{sc}] & 63.2 & 59.4 & 30.8 & 18.0 & 27.4 & 40.6 & 39.9 & \\
        & LLM Debate~[\citenum{debate1}] & 71.2 & \underline{59.6} & 28.8 & 29.4 & 29.8 & \textbf{43.4} & 43.7 & \\
        & Div-MAD~[\citenum{debate2}] & 61.8 & 51.4 & 28.0 & 21.0 & 27.6 & 37.8 & 37.9 & \\
        & DMAD~[\citenum{liu2025breaking}] & \textbf{76.0} & 56.2 & 27.4 & 37.5 & 28.9 & 40.4 & \underline{44.4} & \\
        \cmidrule(l){2-10}
        & \textbf{DREAM} & \underline{73.0} & \textbf{61.2} & \textbf{32.9} & \textbf{41.3} & \textbf{31.4} & \underline{41.6} & \textbf{46.9} & \textbf{\textcolor{green!50!black}{$\uparrow$3.2}} \\
    \bottomrule
    \end{tabular}
\end{table*}

\subsection{DREAM Outperforms Baselines on M3MAD-Bench}
\label{sec:claim_a}

Table~\ref{tab:main_multimodal} shows the results from three small-scale models (with up to 8 billion parameters), one frontier model, and six benchmarks. Instead of just evaluating isolated performance, our analysis explores the structural limitations of current multi-agent frameworks, emphasizing how DREAM effectively addresses these challenges. GPT-4o-mini with Perceiver was excluded because it could not be used.

\textbf{The Structural Failures of Div-MAD, Perceiver, and DwI.} 
Among the evaluated baselines, LLM Debate and DMAD hit a strict performance plateau, while Div-MAD~\citep{debate2}, Perceiver~\citep{HuangBeCollaboration}, and DwI~\citep{FangDEBATEMODELS} exhibit poor performance. Div-MAD consistently ranks as the weakest baseline in our scope (e.g., dropping to a mere 24.5\% on Gemma-3-4B). We attribute this to its early termination mechanism; in complex multimodal tasks, premature halting interrupts discussions before agents can resolve visual ambiguities and reach consensus. Perceiver-based approaches fail due to the inherent spatial compression bottlenecks~\citep{HuangBeCollaboration}, where mapping visual inputs to a fixed-capacity latent space aggressively destroys fine-grained details. Finally, while DwI attempts to improve visual grounding through spatial zooming, it continues to force the model to process these crops at a fixed, static resolution. This rigid reliance on zooming without dynamic resolution scaling leaves agents unable to effectively balance dense local details with global image context.

\begin{table*}[t]
    \centering
    \caption{Performance of the evaluated models under different wrapper methods (\textit{Perceiver} vs.\ \textit{DREAM}). Best results are \textbf{bolded}, second-best \underline{underlined}. $\Delta$ denotes the average accuracy improvement of DREAM over the corresponding bare debate baseline. \textit{MVt}: MathVista, \textit{MVs}: MathVision, \textit{PVQA}: PathVQA, \textit{MME-R}: MME-Reasoning, \textit{VP}: VisualPuzzles.}
    \label{tab:main_multimodal_generalize}
    \footnotesize
    \setlength{\tabcolsep}{6pt}
    \begin{tabular}{c l c cc c cc c c}
    \toprule
        \multirow{2}{*}{\textbf{Model}} &
        \multirow{2}{*}{\textbf{Method}} &
        \textbf{Know.} &
        \multicolumn{2}{c}{\textbf{Mathematics}} &
        \textbf{Med.} &
        \multicolumn{2}{c}{\textbf{Reasoning}} &
        \multirow{2}{*}{\textbf{Avg.}} &
        \multirow{2}{*}{$\mathbf{\Delta}$} \\
        \cmidrule(lr){3-3} \cmidrule(lr){4-5} \cmidrule(lr){6-6} \cmidrule(lr){7-8}
        & & \textbf{MME} & \textbf{MVt} & \textbf{MVs} & \textbf{PVQA} & \textbf{MME-R} & \textbf{VP} & & \\
    \midrule
    \multirow{8}{*}{\rotatebox[origin=c]{90}{\textcolor{cyan!80!black}{\textbf{Gemma-3-4B}}}}
        & \multicolumn{9}{l}{Perceiver~[\citenum{HuangBeCollaboration}]} \\
        & \quad w/ LLM Debate~[\citenum{debate1}] & \textbf{82.5} & 52.4 & \underline{23.1} & 35.0 & \underline{26.0} & 32.7 & 41.9 & \\
        & \quad w/ Div-MAD~[\citenum{debate2}] & 43.0 & 45.5 & 20.6 & 18.3 & 20.5 & 30.3 & 29.7 & \\
        & \quad w/ DMAD~[\citenum{liu2025breaking}] & \underline{81.7} & \underline{56.1} & 22.1 & 38.9 & 25.8 & 30.0 & 42.4 & \\
        \cmidrule(l){2-10}
        & \multicolumn{9}{l}{\textbf{DREAM}} \\
        & \quad w/ LLM Debate~[\citenum{debate1}] & 80.7 & \textbf{57.3} & \textbf{27.3} & \underline{41.6} & \textbf{28.2} & \textbf{37.0} & \textbf{45.4} & \textbf{\textcolor{green!50!black}{$\uparrow$1.5}} \\
        & \quad w/ Div-MAD~[\citenum{debate2}] & 33.4 & 39.5 & 18.4 & 18.0 & 20.6 & 26.1 & 26.0 & \textbf{\textcolor{green!50!black}{$\uparrow$1.5}} \\
        & \quad w/ DMAD~[\citenum{liu2025breaking}] & 80.1 & 52.3 & 22.9 & \textbf{43.0} & 25.7 & \underline{36.9} & \underline{43.5} & \textbf{\textcolor{green!50!black}{$\uparrow$1.2}} \\
    \midrule
    \multirow{8}{*}{\rotatebox[origin=c]{90}{\textcolor{orange!80!black}{\textbf{Qwen2.5-VL-7B}}}}
        & \multicolumn{9}{l}{Perceiver~[\citenum{HuangBeCollaboration}]} \\
        & \quad w/ LLM Debate~[\citenum{debate1}] & 54.1 & 65.4 & 24.4 & 39.1 & 30.4 & \underline{38.9} & 42.1 & \\
        & \quad w/ Div-MAD~[\citenum{debate2}] & 52.3 & 63.1 & 21.9 & 15.5 & 25.7 & 32.5 & 35.2 & \\
        & \quad w/ DMAD~[\citenum{liu2025breaking}] & 39.3 & 53.5 & 27.8 & 28.0 & 28.5 & 36.1 & 35.5 & \\
        \cmidrule(l){2-10}
        & \multicolumn{9}{l}{\textbf{DREAM}} \\
        & \quad w/ LLM Debate~[\citenum{debate1}] & \textbf{81.9} & \textbf{73.3} & \underline{28.5} & \underline{41.6} & \textbf{34.5} & \textbf{41.6} & \textbf{50.2} & \textbf{\textcolor{green!50!black}{$\uparrow$3.2}} \\
        & \quad w/ Div-MAD~[\citenum{debate2}] & 59.1 & 63.5 & 25.8 & 18.6 & 27.6 & 33.9 & 38.1 & \textbf{\textcolor{green!50!black}{$\uparrow$1.3}} \\
        & \quad w/ DMAD~[\citenum{liu2025breaking}] & \underline{76.7} & \underline{71.6} & \textbf{29.9} & \textbf{47.9} & \underline{33.4} & 37.3 & \underline{49.5} & \textbf{\textcolor{green!50!black}{$\uparrow$2.2}} \\
    \midrule
    \multirow{8}{*}{\rotatebox[origin=c]{90}{\textcolor{violet!80!black}{\textbf{InternVL3-8B}}}}
        & \multicolumn{9}{l}{Perceiver~[\citenum{HuangBeCollaboration}]} \\
        & \quad w/ LLM Debate~[\citenum{debate1}] & 84.8 & 70.0 & 25.5 & 39.1 & 30.7 & 34.7 & 47.5 & \\
        & \quad w/ Div-MAD~[\citenum{debate2}] & 58.9 & 46.1 & 21.5 & 34.2 & 26.3 & 31.7 & 36.5 & \\
        & \quad w/ DMAD~[\citenum{liu2025breaking}] & 80.0 & 64.4 & 23.1 & 41.3 & 26.7 & 35.2 & 45.1 & \\
        \cmidrule(l){2-10}
        & \multicolumn{9}{l}{\textbf{DREAM}} \\
        & \quad w/ LLM Debate~[\citenum{debate1}] & \textbf{89.6} & \textbf{73.3} & 29.4 & \underline{42.9} & \textbf{31.6} & \textbf{41.7} & \textbf{51.4} & \textbf{\textcolor{green!50!black}{$\uparrow$2.5}} \\
        & \quad w/ Div-MAD~[\citenum{debate2}] & 43.4 & 45.1 & \underline{30.7} & 33.9 & 28.0 & 34.0 & 35.9 & \textbf{\textcolor{green!50!black}{$\uparrow$1.5}} \\
        & \quad w/ DMAD~[\citenum{liu2025breaking}] & \underline{85.9} & \underline{70.3} & \textbf{31.9} & \textbf{44.9} & \underline{31.4} & \underline{38.1} & \underline{50.4} & \textbf{\textcolor{green!50!black}{$\uparrow$1.9}} \\
    \midrule
    \multirow{4}{*}{\rotatebox[origin=c]{90}{\textcolor{green!70!black}{\textbf{GPT-4o}}}}
    \multirow{4}{*}{\rotatebox[origin=c]{90}{\textcolor{green!70!black}{\textbf{mini}}}}
        & \multicolumn{9}{l}{\textbf{DREAM}} \\
        & \quad w/ LLM Debate~[\citenum{debate1}] & \underline{73.0} & \textbf{61.2} & \textbf{32.9} & \textbf{41.3} & \textbf{31.4} & \underline{41.6} & \textbf{46.9} & \textbf{\textcolor{green!50!black}{$\uparrow$3.2}} \\
        & \quad w/ Div-MAD~[\citenum{debate2}] & 62.3 & 51.9 & \underline{29.0} & 20.9 & 29.3 & 39.5 & 38.8 & \textbf{\textcolor{green!50!black}{$\uparrow$0.9}} \\
        & \quad w/ DMAD~[\citenum{liu2025breaking}] & \textbf{77.8} & \underline{58.2} & 28.8 & \underline{38.6} & \underline{30.9} & \textbf{42.1} & \underline{46.1} & \textbf{\textcolor{green!50!black}{$\uparrow$1.7}} \\
    \bottomrule
    \end{tabular}
\end{table*}

\textbf{DREAM Shows Consistent, Architecture-Robust Generalization.} 
Dynamic Resolution Assignment (DRA) and Uncertainty-guided Rollback Aggregation (URA) directly address these bottlenecks. Across four diverse model families and all three debate frameworks (yielding 12 distinct model$\times$family combinations), equipping the baseline with DREAM improves the average accuracy in \textbf{all 12 evaluated cases}. These results suggest that DREAM is not an overfitted heuristic, but a robust, plug-and-play wrapper that consistently improves visual scaling and uncertainty-aware aggregation across a range of MLLMs.

\subsection{Generalization Across Debate Frameworks}
\label{sec:wrapper_generalization}

To demonstrate that our approach addresses fundamental flaws in multimodal reasoning rather than overfitting to a specific configuration, we apply DREAM as a modular upgrade to three distinct general debate backbones: standard LLM Debate~\citep{debate1}, Div-MAD~\citep{debate2}, and DMAD~\citep{liu2025breaking}. Because DREAM operates entirely as a plug-and-play wrapper, it integrates seamlessly without altering their core messaging logic. As shown in Table~\ref{tab:main_multimodal_generalize}, equipping these frameworks with DREAM yields universal accuracy improvements across all evaluated MLLMs. In contrast, the Perceiver~\citep{HuangBeCollaboration} wrapper, which isolates perception by compressing the image into a fixed number of latent tokens, consistently degrades performance. This confirms that aggressive spatial compression destroys necessary fine-grained details, whereas DREAM preserves better grounding through dynamic resolution scaling.

Crucially, analyzing these combinations reveals how different backbones interact with visual assignment. The standard LLM Debate backbone pairs best with DREAM, achieving the highest absolute averages (e.g., 51.4\% on InternVL3-8B) because its unrestricted, multi-round exchanges allow DREAM's URA to capture the full trajectory of certainty erosion and roll back groupthink effectively. Conversely, Div-MAD remains the weakest backbone; its early termination mechanism prematurely cuts debates short, preventing agents from resolving visual ambiguities. While DREAM still improves Div-MAD over its bare baseline (e.g., $\uparrow$1.3\% on Qwen), it cannot fully reverse the structural damage of halting reasoning too early. Finally, DMAD—which uses adversarial roles to break textual groupthink—receives strong, stackable boosts from DREAM (e.g., reaching 49.5\% on Qwen), proving that our uncertainty-guided visual rescue provides orthogonal benefits even when the underlying backbone already attempts to mitigate consensus errors.

\section{Ablation Study}
\subsection{URA Rescues Certainty-Erosion Cases}

\begin{wraptable}{r}{0.7\textwidth}
    \centering
    \vspace{-\intextsep}
    \hspace*{-.75\columnsep}
    \caption{Ablation of DREAM components and their contribution to average accuracy (\%).}
    \label{tab:aggregation}
    \resizebox{\linewidth}{!}{
        \begin{tabular}{lccc}
        \toprule
        \textbf{Method} & \textbf{Gemma-3-4B} & \textbf{Qwen2.5-VL-7B} & \textbf{InternVL3-8B} \\
        \midrule
        Baseline (LLM Debate) & 43.9 & 47.0 & 48.9 \\
        + DRA (Resolution Probing) & 44.7 (\textbf{\textcolor{green!50!black}{$\uparrow$0.8}}) & 48.3 (\textbf{\textcolor{green!50!black}{$\uparrow$1.3}}) & 49.6 (\textbf{\textcolor{green!50!black}{$\uparrow$0.7}}) \\
        + URA (Adaptive Threshold) & 45.1 (\textbf{\textcolor{green!50!black}{$\uparrow$1.2}}) & 48.9 (\textbf{\textcolor{green!50!black}{$\uparrow$1.9}}) & 50.5 (\textbf{\textcolor{green!50!black}{$\uparrow$1.6}}) \\
        + URA (Rollback) & 45.4 (\textbf{\textcolor{green!50!black}{$\uparrow$1.5}}) & 50.2 (\textbf{\textcolor{green!50!black}{$\uparrow$3.2}}) & 51.4 (\textbf{\textcolor{green!50!black}{$\uparrow$2.5}}) \\
        \bottomrule
        \end{tabular}
    }
\end{wraptable}

Table~\ref{tab:aggregation} ablates the contribution of each DREAM module to the final debate consensus. The addition of DRA consistently improves upon the baseline (e.g., $\uparrow1.3\%$ on Qwen2.5-VL). Implementing URA further compounds these gains; applying the Adaptive Threshold provides a solid intermediate boost, while the complete Confidence-Guided Rescue mechanism achieves peak accuracy across all evaluated models. Crucially, these continuous improvements are observed across fundamentally different model architectures. This demonstrates that DREAM generalizes seamlessly across diverse MLLM families, proving that dynamically rescuing early, high-confidence answers is a universally effective strategy for mitigating late-stage groupthink and hallucination.

To ensure the robustness of these modules, the sensitivity of DREAM's core hyperparameters was systematically evaluated (detailed configurations and full sweep results are provided in Appendix Table~\ref{tab:appendix_hyperparams}). For URA, the rescue sensitivity multiplier $\alpha$ strictly governs the threshold for rollback. Empirical evaluations indicate that $\alpha = 0.5$ yields the optimal balance across all tested models. Lower values trigger rollbacks too aggressively in response to natural token probability variations, thereby disrupting valid late-round consensus. Conversely, higher values (e.g., $\alpha \ge 1.0$) render the rollback mechanism overly conservative, allowing subtle but destructive instances of certainty erosion to slip through uncorrected. Similarly, for DRA, the adaptive thresholding strategy ($\mu_0 - \sigma_0$) systematically outperforms fixed static scalars by dynamically calibrating the confidence boundary to the inherent visual difficulty of each individual query. Appendix~\ref{sec:appendix_branch_analysis} breaks this behavior down further by individual DRA branch, showing that no single branch dominates uniformly across benchmarks and that each trades off accuracy against token cost differently, which motivates assigning per-sample rather than committing to one fixed strategy.

\subsection{Accuracy vs.\ Computation}
\label{sec:pareto}

\begin{wrapfigure}[15]{r}{0.4\textwidth}
    \vspace{-\intextsep}
    \hspace*{-.75\columnsep}
    \includegraphics[width=\linewidth]{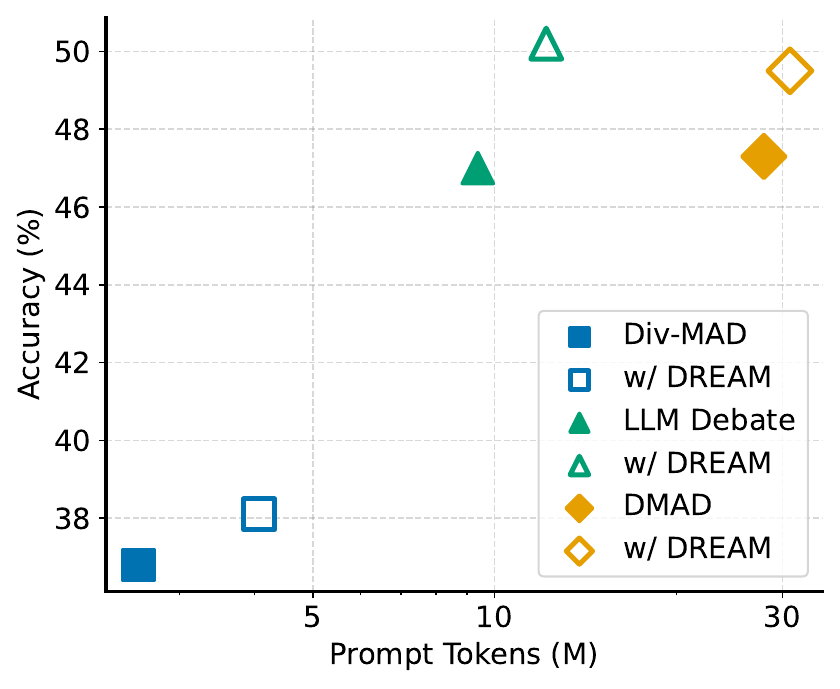}
    \caption{Accuracy vs.\ prompt token consumption for Qwen2.5-VL-7B between baselines and DREAM.}
    \label{fig:pareto}
\end{wrapfigure}

The primary objective of DREAM is not strict token reduction, but rather the optimization of the accuracy-versus-tokens trade-off. Figure~\ref{fig:pareto} illustrates this trade-off by plotting the bare debate baselines against their \textit{w/ DREAM}-wrapped counterparts for Qwen2.5-VL-7B. While accuracy consistently improves across all three debate families (yielding gains of $\uparrow$1.3\% to $\uparrow$3.2\%), this performance is accompanied by an expected increase in token consumption due to multi-scale visual processing. Consequently, the wrapped configurations incur higher computational costs than their bare counterparts. Ultimately, the integration of DREAM systematically shifts the trade-off upward, demonstrating that the strategic investment of additional visual tokens translates directly into superior multimodal reasoning capabilities that standard debate structures cannot achieve.

\section{Conclusion}

We have presented DREAM, a framework designed to overcome critical efficiency and consensus bottlenecks in multi-agent multimodal debate. DREAM introduces two core mechanisms: Dynamic Resolution Assignment (DRA), which transforms image resolution scaling into a data-driven, per-sample decision using zero-shot ANLL uncertainty signals; and Uncertainty-guided Rollback Aggregation (URA), which mitigates debate groupthink by tracking per-agent uncertainty trajectories and rescuing high-confidence answers before they are overridden by group pressure. Across six multimodal reasoning benchmarks, wrapping existing debate baselines with DREAM consistently improves their accuracy-vs-computation trade-off without requiring dataset-specific tuning.
\section*{Reproducibility statement}

The approach is described in Section \ref{sec:method} and provided as pseudocode in Appendix \ref{sec:appendix_algorithm}. Details of the implementation and hyperparameter choices are reported in Appendix \ref{app:implementation}. All experiments use frozen models and publicly available checkpoints, with inference run using three different random seeds. The code and checkpoints will be made available.







\section*{Ethics Statement}

Our work focuses on improving the reasoning accuracy of multimodal AI systems through multi-agent debate. We do not introduce new datasets or collect human data. The benchmarks used in our evaluation are publicly available and widely used in the research community. We acknowledge that improvements in multimodal reasoning capabilities could have both beneficial applications (e.g., medical image analysis, educational tools) and potential risks (e.g., more convincing misinformation generation). We encourage responsible deployment of such systems with appropriate safeguards.

\section*{AI Use Statement}

In this work, we used generative AI tools to assist with manuscript editing. We have not used generative AI tools to formulate the core research ideas, conduct the empirical experiments, or generate the evaluation data. Specifically, all AI-refined text was manually verified by the authors to ensure it perfectly aligns with our scientific claims and empirical findings, and all AI-generated formatting code was tested for compilation correctness. We take full responsibility for the final content of this work, including any text, claims, or artifacts produced with the aid of generative AI.

\bibliography{iclr2027_conference}
\bibliographystyle{iclr2027_conference}

\clearpage
\appendix

\section{Dataset Details}
\label{sec:appendix_datasets}

Full descriptions of the six benchmarks used in evaluation (multimodal evaluation subset of M3MAD-Bench \citep{li2026m3mad}):

\begin{itemize}
    \item \textbf{MME} \citep{mme}: 14 subtasks spanning perception (OCR, color recognition) and cognition (commonsense reasoning).
    \item \textbf{MathVista} \citep{lu2024mathvista}: 6,141 visual mathematical reasoning tasks from 31 source datasets.
    \item \textbf{MathVision} \citep{wang2024mathvision}: 3,040 visual math problems over geometric figures, plots, and diagrams.
    \item \textbf{PathVQA} \citep{pathvqa}: pathology-image VQA requiring microscopic feature identification.
    \item \textbf{MME-Reas.} \citep{mme-reasoning}: abductive, deductive, and analogical reasoning, extending MME.
    \item \textbf{VisualPuzzles} \citep{visualpuzzles}: spatial and abstract visual puzzle reasoning.
\end{itemize}

\section{Implementation Details.}
\label{app:implementation}
To ensure cost efficiency, we randomly sampled 500 examples from each dataset for evaluation, following the approach outlined in M3MAD~\citep{li2026m3mad}. We set the temperature to 0.5 and the maximum token length to 4096. To achieve more accurate answer assessments, we employed a model specifically designed for evaluation, called x-Verify~\citep{xVerify}. This model utilizes an LLM-as-a-judge paradigm to provide scalable and consistent evaluations across various tasks and modalities.

For the methods evaluated, we maintained consistent hyperparameter settings specific to each method (e.g., the number of agents and debate rounds) in accordance with their official implementations. Specifically, the Self-Consistency (SC) method uses \(N = 5\) independent reasoning paths with majority voting. LLM Debate employs \(N = 3\) agents over \(N = 3\) debate rounds. The Div-MAD approach involves two role-based agents ("Affirmative" and "Negative") with \(N = 3\) rounds. Similarly, DMAD also utilizes \(N = 3\) agents across \(N = 3\) rounds. In alignment with the original paper, DMAD assigns different reasoning strategies to its agents to promote diversity.

For unimodal tasks, we employed the following strategies: Chain-of-Thought (CoT) \citep{langchain}, Step-Back Prompting (SBP) \citep{zheng2024take}, and Program-of-Thought (PoT) \citep{chen2023program}. For multimodal tasks, we utilized IO inference, Compositional Chain-of-Thought (CCoT) \citep{mitra2024compositional}, and Duty-Distinct Chain-of-Thought (DDCoT) \cite{zheng2023ddcot}. The experiments were conducted on eight NVIDIA A100 GPUs with 40GB of memory, while access to GPT-4o-mini was facilitated through OpenAI API calls.

\section{Assignment Strategy Definitions}
\label{app:assignment_strategies}

The Dynamic Resolution Assignment (DRA) step selects an image payload for each agent to use across all debate rounds. We define the five named strategies below, reusing the main text's notation: $\mathcal{S} = \{s_1, \ldots, s_K\}$ is the full resolution candidate set, $\mathcal{C}$ the confident agent set after thresholding, $\mathcal{S}^*_\mathcal{C} = \{s_i^{(0)} : i \in \mathcal{C}\}$ the resolutions at which confident agents were probed, and $\mathcal{S}^\dagger \subseteq \mathcal{S}^*_\mathcal{C}$ the majority-agreeing subset (used only in Case~5). A \emph{multi-scale payload} of a set $\mathcal{X}$ means the image is encoded at every resolution in $\mathcal{X}$ and all resulting visual tokens are concatenated into a single input.

\paragraph{\textsc{MixRes}.} All $N$ agents receive the full multi-scale payload $\mathcal{S}$. Applied when $|\mathcal{C}| = 0$ (Case~1): no agent was confident at any resolution, so all scales are provided as a fallback rather than discarding any signal.

\paragraph{\textsc{BestANLL}.} All $N$ agents are locked to the single resolution in $\mathcal{S}^*_\mathcal{C}$ (or, in the Case~5 majority-filter branch below, in $\mathcal{S}^\dagger$). Applied in two situations: (i) exactly one confident agent (Case~2), where $\mathcal{S}^*_\mathcal{C}$ is already a singleton; and (ii) the confident-subset-agrees branch of Case~4 or the majority-filter branch of Case~5, where $|\mathcal{S}^*_\mathcal{C}| = 1$ or $|\mathcal{S}^\dagger| = 1$ collapses the mixture to a single scale.

\paragraph{\textsc{StaticAll}.} Each agent $i$ retains its own probe-round resolution $s_i^{(0)}$; no agent changes scale. Applied when all $N$ agents were confident and in agreement (Case~3), preserving the full resolution diversity that the probe was designed to explore.

\paragraph{\textsc{SubsetStaticMixed}.} Confident agents ($i \in \mathcal{C}$) keep their own probe resolutions $s_i^{(0)}$. Unconfident agents ($i \notin \mathcal{C}$) receive the multi-scale mixture of $\mathcal{S}^*_\mathcal{C}$ (the set of resolutions endorsed by the confident, agreeing subset). Applied when a strict confident subset agrees and spans more than one resolution (Case~4, $|\mathcal{S}^*_\mathcal{C}| > 1$). When $|\mathcal{S}^*_\mathcal{C}| = 1$, \textsc{SubsetStaticMixed} reduces to \textsc{BestANLL}.

\paragraph{\textsc{ConflictMixRes}.} All $N$ agents receive a multi-scale mixture of a resolution subset. Applied in Case~5 (confident agents disagree):
\begin{itemize}
    \item \emph{Strict majority exists} (at least two confident agents share the same answer): dissenting agents are excluded; all agents receive the mixture of
    $\mathcal{S}^\dagger$ (the majority-agreeing resolutions only). When $|\mathcal{S}^\dagger| = 1$ this again reduces to \textsc{BestANLL}.
    \item \emph{No strict majority}: all confident resolutions $\mathcal{S}^*_\mathcal{C}$ are used without filtering, since no subset can be identified as more reliable.
\end{itemize}

\section{Token Efficiency Detail}
\label{sec:appendix_efficiency}

As detailed in Section~\ref{sec:pareto}, DREAM fundamentally improves the accuracy-efficiency trade-off for multimodal multi-agent systems. Rather than optimizing for a naive, uniform reduction in token usage, which typically starves models of critical visual context, DREAM strategically reallocates computational bandwidth to where it is needed most. Table~\ref{tab:efficiency} presents the comprehensive token economics for Qwen2.5-VL-7B (with accuracy figures corresponding to the averages in Table~\ref{tab:main_multimodal}). This breakdown definitively shows that by accepting a marginal token overhead during the initial probing phase, DREAM successfully prevents the massive, compounding token waste caused by hallucination loops in later rounds. Ultimately, DREAM establishes a commanding accuracy-efficiency trade-off, achieving state-of-the-art accuracy through highly intelligent, cost-efficient token assignment.

\begin{table}[!htp]
\centering
\caption{Prompt token consumption and accuracy for Qwen2.5-VL-7B.}
\label{tab:efficiency}
\begin{tabular}{lcc}
\toprule
\textbf{Method} & \textbf{Prompt Tokens (M)} & \textbf{Accuracy (\%)} \\
\midrule
Div-MAD & 2.57 & 36.8 \\
\quad\textit{w/ DREAM} & 4.07 & 38.1 \\
\midrule
LLM Debate & 9.38 & 47.0 \\
\quad\textit{w/ DREAM} & 12.18 & 50.2 \\
\midrule
DMAD & 27.96 & 47.3 \\
\quad\textit{w/ DREAM}  & 30.88 & 49.5 \\
\bottomrule
\end{tabular}
\end{table}

The LLM Debate row's bare-token figure uses the 3-round debate configuration, matching DREAM's own $R=3$ debate budget. Under \textit{w/ DREAM}, all three families improve in accuracy over their own bare row ($\uparrow$1.3 to $\uparrow$3.2), matching Table~\ref{tab:main_multimodal}'s $\Delta$ column, but token consumption also increases for all three, consistent with the added cost of the probing round and, in some cases, the routed resolution set; see Section~\ref{sec:pareto} for the resulting accuracy-vs-tokens trade-off.

\section{Extended Results: Additional Model}
\label{sec:appendix_large_models}

\begin{table*}[!htp]
    \centering
    \caption{Performance on M3MAD-Bench multimodal benchmarks for LLaVA-Next-7B. Best result per column among rows with reported values \textbf{bolded}, second-best \underline{underlined}. $\Delta$ denotes \textit{w/ DREAM}'s average accuracy change over its bare debate baseline. \textit{MVt}: MathVista, \textit{MVs}: MathVision, \textit{PVQA}: PathVQA, \textit{MME-R}: MME-Reasoning, \textit{VP}: VisualPuzzles.}
    \label{tab:appendix_multimodal}
    \footnotesize
    \setlength{\tabcolsep}{6pt}
    \begin{tabular}{c l c cc c cc c c}
    \toprule
        \multirow{2}{*}{\textbf{Model}} &
        \multirow{2}{*}{\textbf{Method}} &
        \textbf{Know.} &
        \multicolumn{2}{c}{\textbf{Mathematics}} &
        \textbf{Med.} &
        \multicolumn{2}{c}{\textbf{Reasoning}} &
        \multirow{2}{*}{\textbf{Avg.}} &
        \multirow{2}{*}{$\mathbf{\Delta}$} \\
        \cmidrule(lr){3-3} \cmidrule(lr){4-5} \cmidrule(lr){6-6} \cmidrule(lr){7-8}
        & & \textbf{MME} & \textbf{MVt} & \textbf{MVs} & \textbf{PVQA} & \textbf{MME-R} & \textbf{VP} &  &  \\
    \midrule
    \multirow{13}{*}{\rotatebox[origin=c]{90}{\textcolor{teal!80!black}{\textbf{LLaVA-Next-7B}}}}
        & IO & 40.2 & 24.0 & 8.0 & 11.2 & 12.8 & 18.0 & 19.0 & \\
        & CoT~[\citenum{wei2022chain}] & 30.2 & 24.0 & 7.2 & 4.6 & 14.8 & 16.0 & 16.1 & \\
        & SC~[\citenum{sc}] & 29.2 & 24.8 & 9.8 & 3.4 & 13.4 & 20.0 & 16.8 & \\
        & DwI~[\citenum{FangDEBATEMODELS}] & 58.9 & 27.2 & 9.3 & 29.5 & 16.0 & 20.8 & 26.9 & \\
        \cmidrule(l){2-10}
        & LLM Debate~[\citenum{debate1}] & \underline{72.0} & \textbf{34.5} & 9.9 & 23.7 & \underline{17.3} & 22.4 & \underline{30.0} & \\
        & \quad\textit{w/ Perceiver~[\citenum{HuangBeCollaboration}]} & 68.0 & 29.5 & 9.8 & 20.9 & 16.3 & 18.9 & 27.2 & \\
        & \quad\textbf{w/ DREAM} & \textbf{72.1} & \underline{34.7} & 10.7 & 25.1 & 17.3 & \textbf{25.1} & \textbf{30.1} & \textbf{\textcolor{green!50!black}{$\uparrow$0.1}} \\
        \cmidrule(l){2-10}
        & Div-MAD~[\citenum{debate2}] & 51.5 & 26.3 & 9.4 & 22.6 & 16.7 & 19.5 & 24.3 & \\
        & \quad\textit{w/ Perceiver~[\citenum{HuangBeCollaboration}]} & 32.7 & 15.4 & 6.5 & 5.6 & 7.3 & 8.1 & 12.6 & \\
        & \quad\textbf{w/ DREAM} & 50.1 & 27.0 & 9.3 & 23.1 & 16.6 & 18.6 & 23.6 & \textbf{\textcolor{red!60!black}{$\downarrow$0.7}} \\
        \cmidrule(l){2-10}
        & DMAD~[\citenum{liu2025breaking}] & 59.3 & 29.5 & \underline{11.4} & \underline{27.1} & 16.3 & 21.9 & 27.6 & \\
        & \quad\textit{w/ Perceiver~[\citenum{HuangBeCollaboration}]} & 57.4 & 26.6 & 8.3 & 21.4 & 16.0 & 16.7 & 24.4 & \\
        & \quad\textbf{w/ DREAM} & 62.3 & 30.7 & \textbf{11.7} & \textbf{28.8} & \textbf{17.8} & \underline{23.1} & 29.1 & \textbf{\textcolor{green!50!black}{$\uparrow$1.5}} \\
    \bottomrule
    \end{tabular}
\end{table*}

Table~\ref{tab:appendix_multimodal} extends our evaluation to LLaVA-Next-7B. Consistent with the near-uniform gains observed on the main-table models, \textit{w/ DREAM} improves two of its three debate families: LLM Debate rises by $\uparrow$0.1\% and DMAD by $\uparrow$1.5\%. Div-MAD is the one exception, regressing by $\downarrow$0.7\%.

\section{Algorithm Pseudocode}
\label{sec:appendix_algorithm}

Algorithm~\ref{alg:DREAM} details the full DREAM procedure, including the DRA and URA phases described in the main methodology.

\begin{algorithm}[tb]
\caption{DREAM: Per-agent Resolution-Informed Scalable Multi-agent debate}
\label{alg:DREAM}
\textbf{Input}: Image $I$, question $q$, resolution set $\mathcal{S}$, agents $\{1, \ldots, N\}$, debate rounds $R$, rescue threshold $\alpha$\\
\textbf{Output}: Final answer $a^*$
\begin{algorithmic}[1]
\STATE \textbf{// Phase 1: Dynamic Resolution Assignment (DRA)}
\FOR{each agent $i \in \{1, \ldots, N\}$}
    \STATE Assign resolution $s_i^{(0)} \in \mathcal{S}$ (round-robin)
    \STATE $\hat{a}_i^{(0)}, \mathrm{ANLL}_i^{(0)} \leftarrow \mathcal{M}(I_{s_i^{(0)}}, q)$
\ENDFOR
\STATE $\tau_{\mathrm{eff}} \leftarrow \mu_0 - \sigma_0$ \hfill \textit{// Adaptive threshold, Eq.~\eqref{eq:threshold}}
\STATE $\mathcal{C} \leftarrow \{i : \mathrm{ANLL}_i^{(0)} \leq \tau_{\mathrm{eff}}\}$ \hfill \textit{// Confident set}
\STATE $\mathcal{S}^* \leftarrow \text{AssignResolutions}(\mathcal{C})$ \hfill \textit{// DRA branch, Appendix~\ref{app:assignment_strategies}}
\STATE \textbf{// Phase 2: Multi-Agent Debate}
\FOR{round $r = 1$ to $R$}
    \FOR{each agent $i \in \{1, \ldots, N\}$}
        \STATE $s_i^{(r)} \leftarrow \mathcal{S}^*[(i-1) \bmod |\mathcal{S}^*|]$ \hfill \textit{// round-robin over routed set}
        \STATE $a_i^{(r)}, \mathrm{ANLL}_i^{(r)} \leftarrow \mathcal{M}(I_{s_i^{(r)}}, q, \text{debate\_context}^{(r-1)})$
    \ENDFOR
\ENDFOR
\STATE \textbf{// Phase 3: Uncertainty-Guided Rollback Aggregation (URA)}
\FOR{each agent $i$}
    \STATE $\sigma_i^{\mathrm{debate}} \leftarrow \text{std}(\{\mathrm{ANLL}_i^{(r)}\}_{r=1}^{R})$ \hfill \textit{// Eq.~\eqref{eq:sigma_debate}}
    \STATE $\delta_i \leftarrow \alpha \cdot \sigma_i^{\mathrm{debate}}$
    \STATE $r_i^* \leftarrow \arg\min_{r < R} \mathrm{ANLL}_i^{(r)}$ \hfill \textit{// Most confident round before the final}
    \IF{$\mathrm{ANLL}_i^{(R)} - \mathrm{ANLL}_i^{(r_i^*)} > \delta_i$}
        \STATE $\hat{a}_i \leftarrow a_i^{(r_i^*)}$ \hfill \textit{// Rescue answer, Eq.~\eqref{eq:rescue}}
    \ELSE
        \STATE $\hat{a}_i \leftarrow a_i^{(R)}$ \hfill \textit{// Keep final answer}
    \ENDIF
\ENDFOR
\STATE $a^* \leftarrow \text{JudgeWithAntiRecency}(\{\hat{a}_i\}, \text{transcript})$
\STATE \textbf{return} $a^*$
\end{algorithmic}
\end{algorithm}

\section{Extended Ablations}
\label{sec:appendix_ablations}

This appendix collects two secondary ablations referenced from the main text: the resolution-scale ablation and the uncertainty-threshold ablation.

\paragraph{Resolution-scale ablation.}

\begin{table}[!htp]
\centering
\caption{Resolution scale ablation (Accuracy \%). $^\dagger$4$\times$ Scale is not stable.}
\label{tab:res_ablation}
\begin{tabular}{lcccc}
\toprule
\textbf{Resolution Strategy} & \textbf{InternVL3-14B} & \textbf{InternVL3-8B} & \textbf{Qwen2.5-VL-7B} & \textbf{Average} \\
\midrule
0.5$\times$ Scale & 49.84 & 49.93 & \textbf{49.45} & 49.74 \\
2$\times$ Scale & 49.78 & 49.48 & 49.31 & 49.52 \\
4$\times$ Scale & \textbf{50.24} & \textbf{50.35}$^\dagger$ & 49.23 & \textbf{49.94}$^\dagger$ \\
\bottomrule
\end{tabular}
\end{table}

Table~\ref{tab:res_ablation} compares fixed-resolution policies applied to the same DMAD + URA backbone: committing every agent to a single resolution for the whole debate, rather than dynamically assigning resolution per agent. We exclude the $4\times$-Scale row from the headline comparison as it frequently encounters maximum token overflow, resulting in truncated context and failed generations, meaning its nominal average is not a reliable comparison point. Excluding it, \textit{$0.5\times$} is the stronger fixed policy on all three models. This ablation isolates the cost of committing to one resolution in advance; it does not include DREAM's full dynamic-assignment configuration, which is evaluated against the bare debate baselines directly in Table~\ref{tab:main_multimodal}.

\paragraph{Uncertainty-threshold ablation.}

\begin{table}[!htp]
\centering
\caption{Ablation on uncertainty thresholds across models (Accuracy \%). $\Delta$ is each row's average minus the best static-threshold average in hindsight (49.74, achieved at 0.20).}
\label{tab:probe_ablation}
\begin{tabular}{lccccc}
\toprule
\textbf{Threshold} & \textbf{InternVL3-14B} & \textbf{InternVL3-8B} & \textbf{Qwen2.5-VL-7B} & \textbf{Avg.} & \textbf{$\mathbf{\Delta}$} \\
\midrule
Static 0.10 & 50.25 & 49.66 & \underline{49.29} & 49.73 & -0.01 \\
Static 0.20 & 50.12 & \underline{50.16} & 48.93 & \underline{49.74} & 0.00 \\
Static 0.30 & \underline{50.51} & 49.57 & 48.90 & 49.66 & \textbf{\textcolor{red!50!black}{$\downarrow$0.08}} \\
Static 0.50 & 50.36 & 49.60 & 49.05 & 49.67 & \textbf{\textcolor{red!50!black}{$\downarrow$0.07}} \\
\textbf{Adaptive ($\mu{-}\sigma$)} & \textbf{50.70} & \textbf{50.40} & \textbf{49.50} & \textbf{50.20} & \textbf{\textcolor{green!50!black}{$\uparrow$0.46}} \\
\bottomrule
\end{tabular}
\end{table}

Table~\ref{tab:probe_ablation} compares static thresholds against our adaptive mechanism. Notably, the adaptive rule does not merely act as a convenient fallback; it strictly outperforms fixed thresholds. No single static threshold is optimal across all three models (e.g., 0.10 wins on the 7B, 0.20 wins on the 8B, and 0.30 wins on the 14B). However, the adaptive rule dynamically adjusts to the per-sample uncertainty distribution, yielding higher accuracy on \textit{every individual model} than even a ``tuned oracle'' that picks the best static threshold in hindsight. This demonstrates that our data-driven thresholding is a strict performance upgrade over static hyperparameter tuning.

\subsection{Hyperparameters Study}
\label{sec:appendix_hyperparams}

\paragraph{Rescue-multiplier ($\alpha$) sweep.}

\begin{table}[!htp]
\centering
\caption{URA rescue-multiplier ($\alpha$) sensitivity sweep (Accuracy \%, Qwen2.5-VL-7B).}
\label{tab:appendix_hyperparams}
\begin{tabular}{lc}
\toprule
\textbf{Rescue Multiplier ($\alpha$)} & \textbf{Qwen2.5-VL-7B} \\
\midrule
0.10 & 42.2 \\
0.25 & 43.3 \\
\textbf{0.50} & \textbf{50.2} \\
0.75 & 47.2 \\
1.00 & 45.3 \\
\bottomrule
\end{tabular}
\end{table}

Table~\ref{tab:appendix_hyperparams} sweeps the rescue multiplier $\alpha$, which sets the per-agent rollback threshold $\delta_i = \alpha \cdot \sigma_i^{\mathrm{debate}}$ in Eq.~(5). Accuracy rises sharply from $\alpha=0.10$ to $\alpha=0.50$ and falls off past that point: low $\alpha$ narrows the threshold so much that ordinary round-to-round fluctuation in token probabilities triggers a rollback, discarding valid late-round consensus, while $\alpha \geq 0.75$ widens the threshold enough that genuine certainty erosion no longer crosses it and goes uncorrected. $\alpha=0.5$ is the peak on this model, consistent with its use as the fixed value across all reported results.

\subsection{Branch-Level Analysis of Dynamic Resolution Assignment}
\label{sec:appendix_branch_analysis}

To characterize how the five DRA branches defined in Appendix~\ref{app:assignment_strategies} (\textsc{MixRes}, \textsc{BestANLL}, \textsc{StaticAll}, \textsc{SubsetStaticMixed}, \textsc{ConflictMixRes}) actually behave in practice, we break down their firing frequency, accuracy, and token cost on Qwen2.5-VL-7B. The pooled DRA accuracy reported below (50.2\%) reproduces the $\alpha=0.5$ row of Table~\ref{tab:appendix_hyperparams}, confirming both figures draw from the same run.

\begin{figure}[!htp]
\centering
\includegraphics[width=\linewidth]{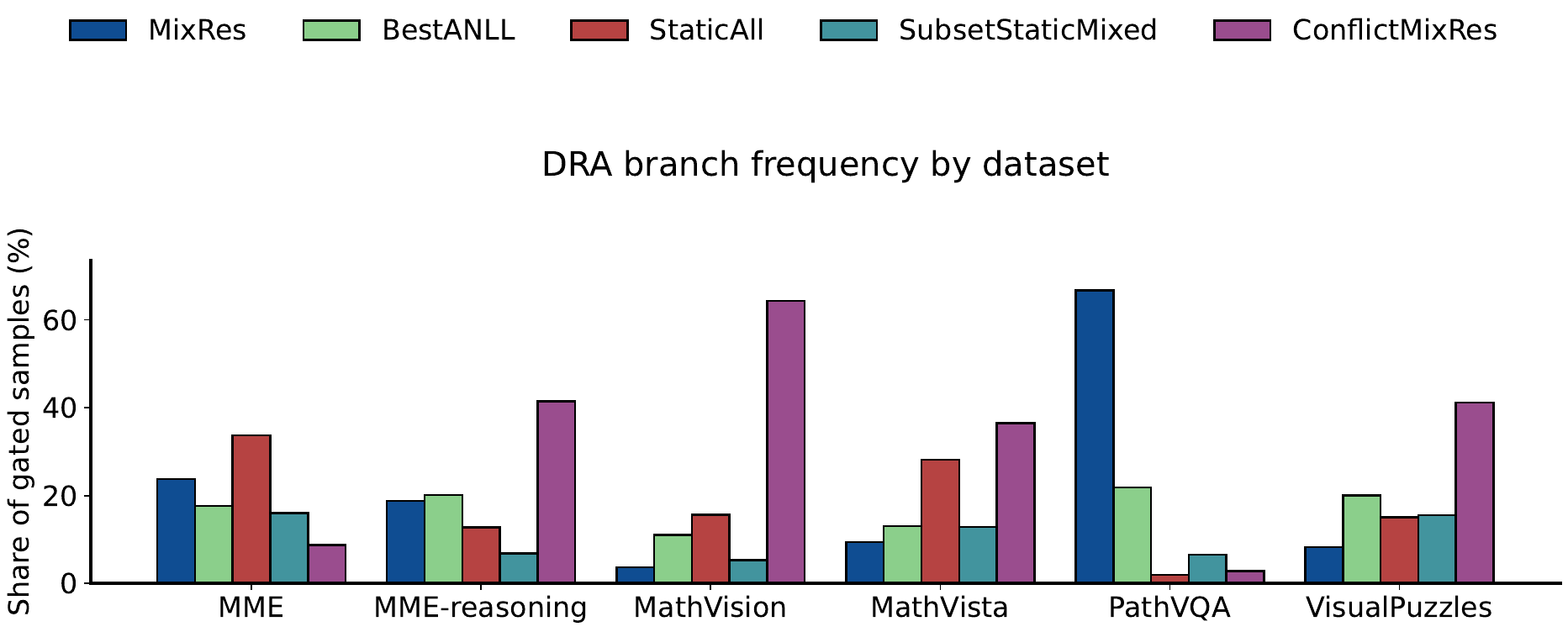}
\caption{Share of gated samples routed to each DRA branch, by benchmark. \textsc{ConflictMixRes} dominates on MathVision (64\%), MME-Reasoning (41\%), and VisualPuzzles (41\%), while \textsc{MixRes} dominates on PathVQA (67\%); \textsc{StaticAll} is comparatively more common on MME (34\%) and MathVista (28\%). No single branch dominates uniformly across benchmarks, consistent with the probe resolving a genuinely per-sample, per-dataset routing decision rather than converging to one default behavior.}
\label{fig:dra_branch_frequency}
\end{figure}

\begin{figure}[!htp]
\centering
\includegraphics[width=\linewidth]{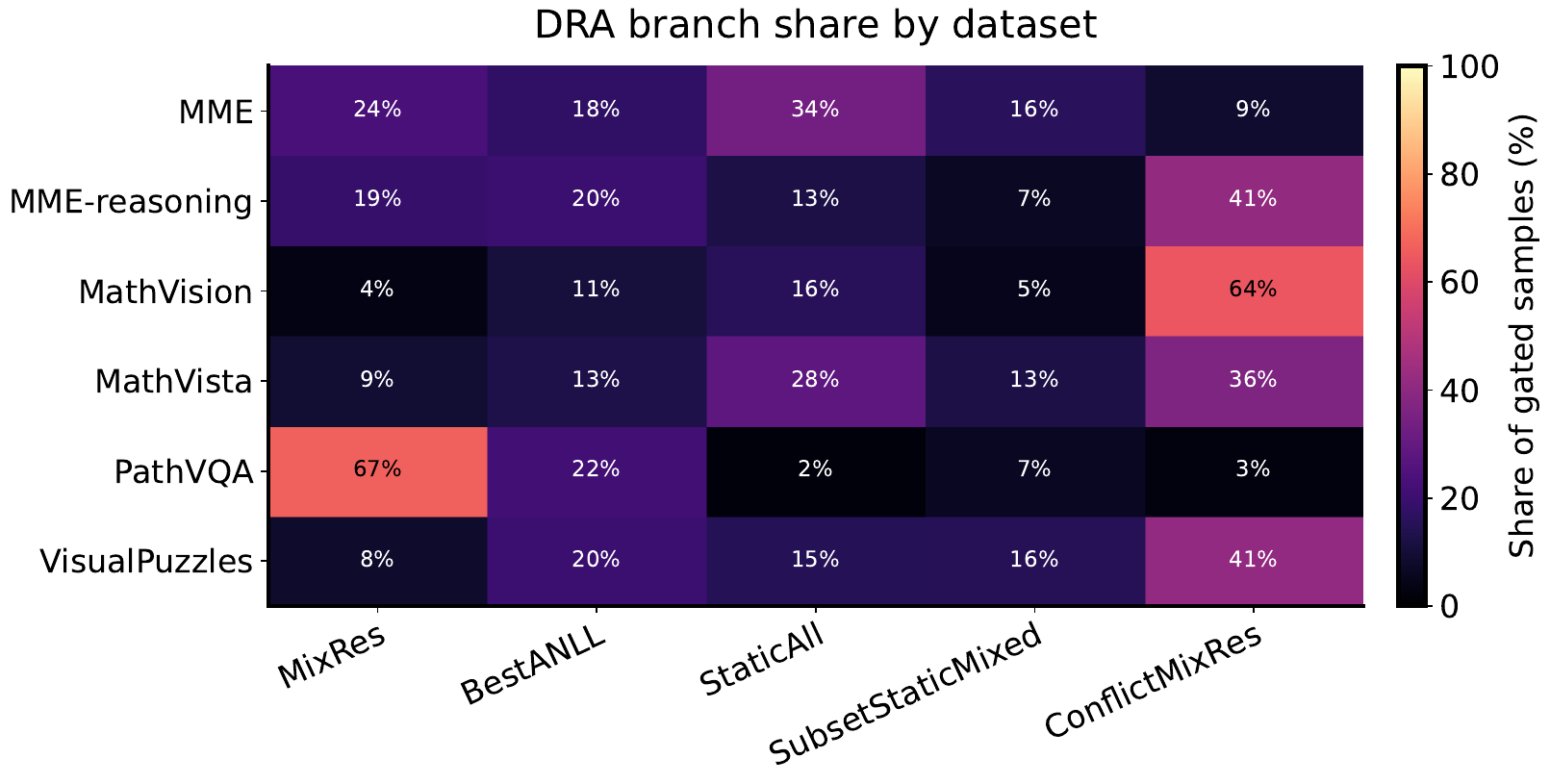}
\caption{Same branch-share data as Figure~\ref{fig:dra_branch_frequency}, shown as a benchmark-by-branch heatmap for direct cross-dataset comparison.}
\label{fig:dra_branch_share_heatmap}
\end{figure}

\begin{figure}[!htp]
\centering
\includegraphics[width=\linewidth]{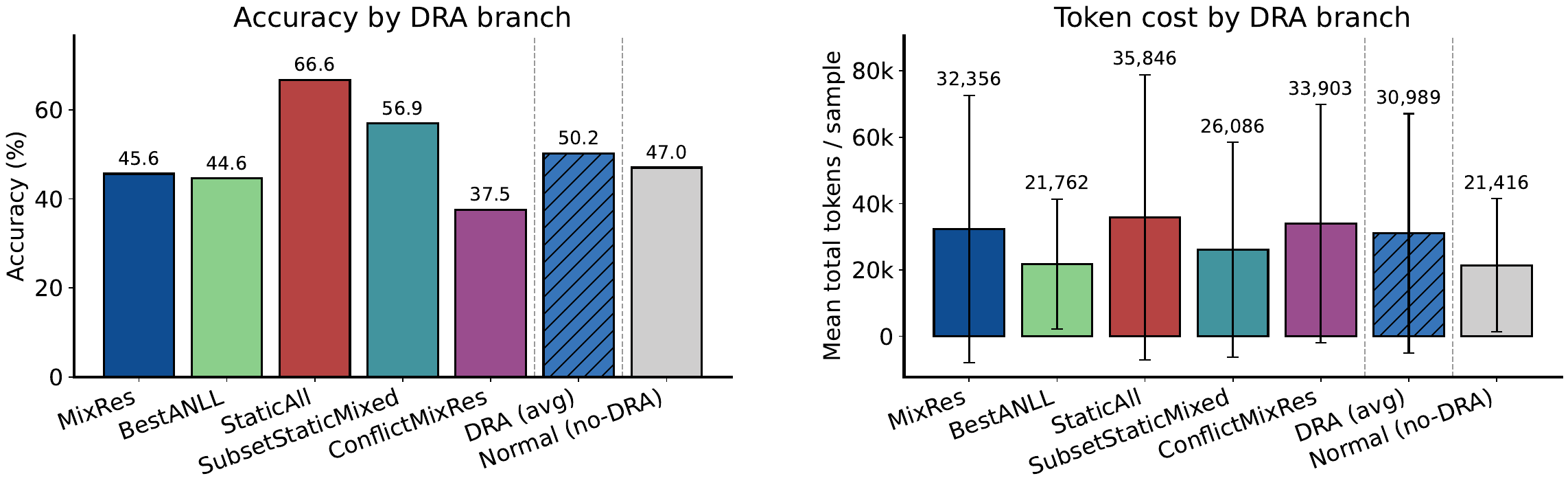}
\caption{Accuracy (left) and mean token cost (right) per DRA branch, pooled across all six benchmarks, against the DRA-average and the no-DRA baseline. \textsc{StaticAll} is the most accurate individual branch (66.6\%) but also among the most expensive (35,846 tokens/sample); \textsc{ConflictMixRes} is the least accurate (37.5\%) despite a high token cost (33,903), reflecting that it fires precisely when confident agents disagree and no majority resolution can be trusted. The DRA-average (50.2\%, 30,989 tokens) exceeds the no-DRA baseline (47.0\%, 21,416 tokens) in accuracy at roughly $1.45\times$ the token cost.}
\label{fig:dra_branch_accuracy_cost}
\end{figure}

\begin{figure}[!htp]
\centering
\includegraphics[width=\linewidth]{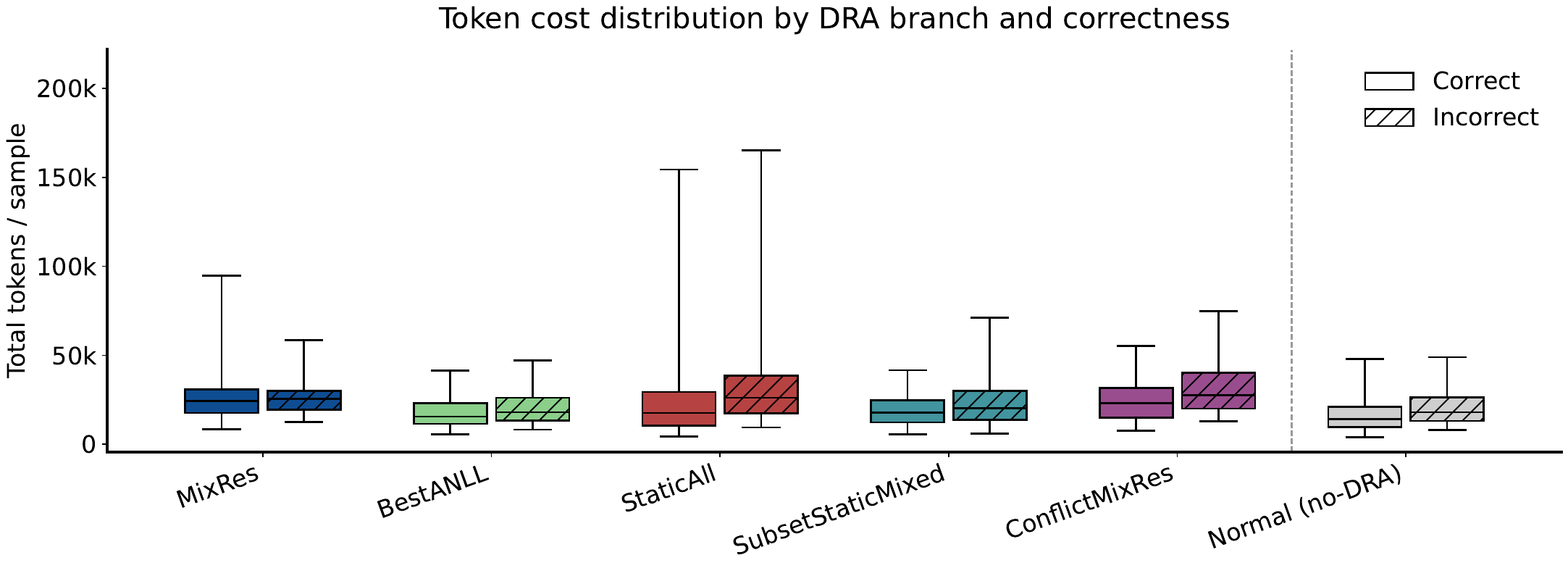}
\caption{Distribution of total tokens per sample for each DRA branch, split by whether the final answer was correct. Within most branches, incorrect samples show a higher median and a longer upper tail than correct samples, most visibly for \textsc{StaticAll}, whose incorrect-sample whisker extends past 165k tokens; the no-DRA baseline shows the same correct/incorrect gap at a lower absolute cost.}
\label{fig:dra_branch_token_distribution}
\end{figure}

Together, these four views indicate that the DRA branches are not interchangeable: they fire under different diagnostic conditions (Appendix~\ref{app:assignment_strategies}), at different rates per benchmark, and trade off accuracy against token cost differently. The pattern that incorrect answers cost more tokens than correct ones within most branches, including the no-DRA baseline, suggests token overrun is itself a symptom of the model struggling on a sample rather than a cause distinct to any one branch.




\section{Case Studies}
\label{sec:appendix_case_studies}

\subsection{Qualitative Analysis}
\label{sec:qualitative_analysis}

\paragraph{Mitigating Visual Hallucinations.} 
In Figure~\ref{fig:qual_mme}, we observe a catastrophic failure mode of standard multi-agent debate on the MME benchmark. Bound to a static resolution, the baseline agents lose grounding and enter a compounding hallucination loop, generating verbose descriptions detailing non-existent ``cosmos flowers'' and a ``bumblebee'' rather than simply identifying the Ringkirche. Conversely, DREAM's Dynamic Resolution Assignment (DRA) probes the image and identifies $2.0\times$ as the optimal, lowest-uncertainty scale (ANLL: $0.106$). By aggressively assigning the debate to this single optimal resolution (\textsc{ConsensusScaling}), DREAM completely suppresses the hallucination loop and returns the correct answer with high precision.

\begin{figure*}[t]
\centering
\begin{tcolorbox}[colframe=black!70, colback=white, boxrule=1pt, arc=4pt, title=\textbf{Case Study: Breaking Sycophancy (MathVision)} - \textit{Query:} Which sparrow has turned around? (GT: \textbf{B: Bernhard})]

\begin{center}
\includegraphics[width=0.4\textwidth, height=4cm, keepaspectratio]{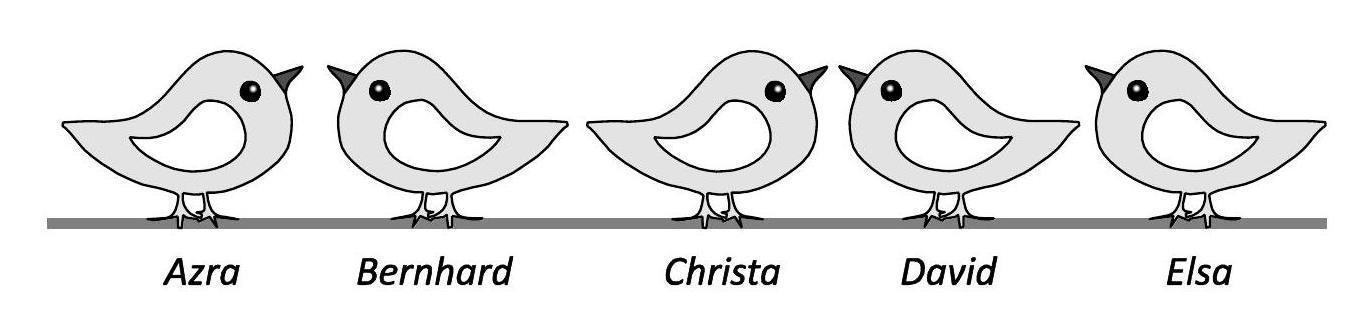}
\end{center}
\vspace{0.2cm}

\begin{minipage}[t]{0.48\textwidth}
\centering
\textbf{\large Standard MAD Baseline} \\
\vspace{0.2cm}
\textit{Fixed Resolution $\rightarrow$ Sycophancy \& Incorrect Logic} \\
\vspace{0.2cm}
\begin{tcolorbox}[colframe=red!60!black, colback=red!5, boxrule=0.5pt, arc=2pt, title=\textbf{Debate (Fixed 1.0x)}]

\footnotesize
\textbf{Agent 1:} ``...This happens when the sparrow was originally at the end of the line (Elsa).'' \\
\textbf{Agent 2:} ``The argument suggests... Based on this logic, the sparrow who turned around should be the one at the end, which is Elsa.'' \\
\vspace{0.1cm}
\textcolor{red!80!black}{\ding{55} Blind Sycophancy (Groupthink)}
\end{tcolorbox}
\vspace{0.2cm}
\textbf{Final Consensus:} \textcolor{red!80!black}{\ding{55} Incorrect (Elsa)}
\end{minipage}%
\hfill\vrule\hfill
\begin{minipage}[t]{0.48\textwidth}
\centering
\textbf{\large DREAM (Ours)} \\
\vspace{0.2cm}
\textit{Dissent-Filtered assignment $\rightarrow$ Accurate Logic} \\
\vspace{0.2cm}
\begin{tcolorbox}[colframe=blue!60!black, colback=blue!5, boxrule=0.5pt, arc=2pt, title=\textbf{Round 0: DRA Probing (\texttt{conflict\_mix\_res})}]

\footnotesize
Probe 0.5x $\rightarrow$ \textbf{D} \quad|\quad Probe 1.0x $\rightarrow$ \textbf{B} \quad|\quad Probe 2.0x $\rightarrow$ \textbf{C} \\
\vspace{0.1cm}
\textit{Decision: Disagreement Detected. Triggering mixed-resolution payload ($2.0\times, 1.0\times, 0.5\times$)}
\end{tcolorbox}
\vspace{-0.1cm}
\begin{tcolorbox}[colframe=green!60!black, colback=green!5, boxrule=0.5pt, arc=2pt, title=\textbf{Debate (Mixed Res Payload)}]

\footnotesize
\textbf{Agents:} ``Let's analyze the situation step by step... Based on the detailed analysis provided, the sparrow that turned around to make the total whistles increase correctly points to Bernhard.'' \\
\vspace{0.1cm}
\textcolor{green!60!black}{\ding{51} Multi-Scale Verification Succeeds}
\end{tcolorbox}
\vspace{0.2cm}
\textbf{Final Consensus:} \textcolor{green!60!black}{\ding{51} Correct ((B) Bernhard)}
\end{minipage}
\end{tcolorbox}
\caption{\textbf{Dismantling Groupthink.} Qualitative comparison on MathVision. The standard baseline succumbs to agreement bias, where subsequent agents blindly agree with an initial flawed assumption. DREAM detects early disagreement during probing and deploys a mixed-resolution payload, forcing the agents to successfully ground their spatial reasoning rather than relying on unverified assumptions.}
\label{fig:qual_mathvision}
\end{figure*}

\paragraph{Dismantling Sycophancy and Groupthink.} 
Figure~\ref{fig:qual_mathvision} highlights DREAM's effectiveness in dismantling sycophancy-a notorious vulnerability in standard multi-agent debate configurations. On a dense spatial reasoning task from MathVision, an early baseline agent hallucinates a flawed logical trajectory, which subsequent agents blindly validate without independent visual verification. DREAM intercepts this failure mode during its Round 0 probing phase. Upon detecting conflicting answers across the independent resolution probes (D, B, and C), the DRA module flags the query's ambiguity and triggers a \textsc{ConflictMixRes} payload. By forcing the agents to reconcile distinct, multi-scale perspectives ($0.5\times$, $1.0\times$, and $2.0\times$), DREAM effectively breaks the groupthink, enabling the agents to systematically verify the spatial layout and converge on the correct answer.

\begin{figure*}[t]
\centering
\begin{tcolorbox}[colframe=black!70, colback=white, boxrule=1pt, arc=4pt, title=\textbf{Case Study: Multi-Scale Clinical Synthesis (PathVQA)} - \textit{Query:} what burn? (GT: \textbf{thermal})]

\begin{center}
\includegraphics[width=0.4\textwidth, height=4cm, keepaspectratio]{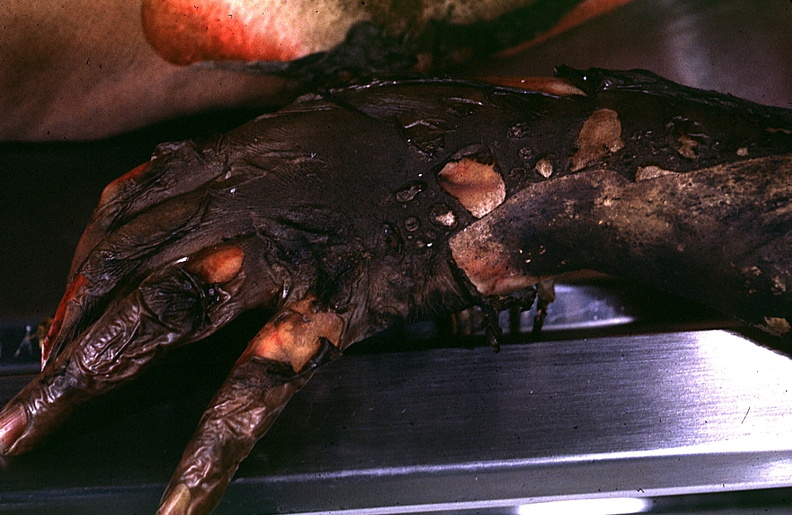}
\end{center}
\vspace{0.2cm}

\begin{minipage}[t]{0.48\textwidth}
\centering
\textbf{\large Standard MAD Baseline} \\
\vspace{0.2cm}
\textit{Fixed Resolution $\rightarrow$ Suboptimal Synthesis} \\
\vspace{0.2cm}
\begin{tcolorbox}[colframe=red!60!black, colback=red!5, boxrule=0.5pt, arc=2pt, title=\textbf{Debate (Fixed 1.0x)}]

\footnotesize
``Based on the descriptions provided, the burn in the image appears to be a severe one... The skin is blackened and charred... indicate that this is a full-thickness burn, most likely a third-degree burn.'' \\
\vspace{0.1cm}
\textcolor{red!80!black}{\ding{55} Static Context Processing}
\end{tcolorbox}
\vspace{0.2cm}
\textbf{Final Consensus:} \textcolor{red!80!black}{\ding{55} Generic Diagnosis}
\end{minipage}%
\hfill\vrule\hfill
\begin{minipage}[t]{0.48\textwidth}
\centering
\textbf{\large DREAM (Ours)} \\
\vspace{0.2cm}
\textit{Multi-Scale Probing $\rightarrow$ Robust Clinical Synthesis} \\
\vspace{0.2cm}
\begin{tcolorbox}[colframe=blue!60!black, colback=blue!5, boxrule=0.5pt, arc=2pt, title=\textbf{Round 0: DRA Probing (\texttt{mix\_res})}]

\footnotesize
Probe 0.5x $\rightarrow$ ANLL: 0.529 (Identifies ``\textbf{thermal injury}'') \\
Probe 1.0x $\rightarrow$ ANLL: 0.547 \quad|\quad Probe 1.5x $\rightarrow$ ANLL: 0.599 \\
\vspace{0.1cm}
\textit{Decision: Complex clinical query. Trigger mixed-resolution assignment.}
\end{tcolorbox}
\vspace{-0.1cm}
\begin{tcolorbox}[colframe=green!60!black, colback=green!5, boxrule=0.5pt, arc=2pt, title=\textbf{Debate (Mixed Res Payload)}]

\footnotesize
\textbf{Agents:} ``The skin is discolored, primarily in shades of black and dark brown, which are indicative of severe \textbf{thermal injury}... highly likely that these are third-degree burns.'' \\
\vspace{0.1cm}
\textcolor{green!60!black}{\ding{51} Comprehensive Clinical Context}
\end{tcolorbox}
\vspace{0.2cm}
\textbf{Final Consensus:} \textcolor{green!60!black}{\ding{51} Precise Etiology Identified}
\end{minipage}
\end{tcolorbox}
\caption{\textbf{Robust Clinical Synthesis.} Qualitative comparison on PathVQA. While the fixed-resolution baseline struggles to extract specific etiology (providing a generic "third-degree" diagnosis), DREAM's multi-scale probing seamlessly extracts complementary clinical cues across distinct visual resolutions. The resulting mixed-resolution debate yields a far more precise and comprehensive medical evaluation.}
\label{fig:qual_pathvqa}
\end{figure*}

\paragraph{Robust Multi-Scale Clinical Synthesis.} 
The advantages of mixed-resolution reasoning extend beyond breaking groupthink, proving highly beneficial for dense domain-specific tasks. As shown in Figure~\ref{fig:qual_pathvqa} (PathVQA), processing a severe burn image at a single fixed resolution limits the baseline's ability to synthesize macro-level tissue damage with micro-level textural cues, resulting in a generic and suboptimal diagnosis. DREAM's DRA probing phase processes the image across multiple scales simultaneously, allowing the $0.5\times$ probe to successfully identify the specific etiology (``thermal injury''). Recognizing the complexity of the medical query via its uncertainty scores, DREAM routes the debate to a \textsc{MixRes} strategy. This allows the debate to seamlessly integrate distinct clinical cues extracted across all three scales, culminating in a highly precise and comprehensive medical evaluation.

\begin{figure*}[t]
\centering
\begin{tcolorbox}[colframe=black!70, colback=white, boxrule=1pt, arc=4pt, title=\textbf{Case Study: Resolving Visual Hallucinations (MME)} - \textit{Query:} Is this a photo of Ringkirche? (GT: \textbf{Yes})]

\begin{center}
\includegraphics[width=0.4\textwidth, height=4cm, keepaspectratio]{figures/0273f127895337be.jpg}
\end{center}
\vspace{0.2cm}

\begin{minipage}[t]{0.48\textwidth}
\centering
\textbf{\large Standard MAD Baseline} \\
\vspace{0.2cm}
\textit{Fixed Resolution $\rightarrow$ Hallucination \& Herding} \\
\vspace{0.2cm}
\begin{tcolorbox}[colframe=red!60!black, colback=red!5, boxrule=0.5pt, arc=2pt, title=\textbf{Debate (Fixed 1.0x)}]

\footnotesize
``Okay, here's a detailed description of the image: The image is a close-up shot of a vibrant garden scene, focusing on a cluster of pink cosmos flowers and a busy bumblebee...'' \\
\vspace{0.1cm}
\textcolor{red!80!black}{\ding{55} Severe Visual Hallucination}
\end{tcolorbox}
\vspace{0.2cm}
\textbf{Final Consensus:} \textcolor{red!80!black}{\ding{55} Incorrect (Failed to answer)}
\end{minipage}%
\hfill\vrule\hfill
\begin{minipage}[t]{0.48\textwidth}
\centering
\textbf{\large DREAM (Ours)} \\
\vspace{0.2cm}
\textit{Dynamic Resolution $\rightarrow$ Accurate Consensus} \\
\vspace{0.2cm}
\begin{tcolorbox}[colframe=blue!60!black, colback=blue!5, boxrule=0.5pt, arc=2pt, title=\textbf{Round 0: DRA Probing (\texttt{consensus\_scaling})}]

\footnotesize
Probe 0.5x $\rightarrow$ ANLL: 0.236 \quad|\quad Probe 1.0x $\rightarrow$ ANLL: 0.258 \\
Probe 2.0x $\rightarrow$ \textbf{ANLL: 0.106 (Lowest Uncertainty)} \\
\vspace{0.1cm}
\textit{Decision: Route debate to optimal \textbf{2.0x} resolution.}
\end{tcolorbox}
\vspace{-0.1cm}
\begin{tcolorbox}[colframe=green!60!black, colback=green!5, boxrule=0.5pt, arc=2pt, title=\textbf{Debate at 2.0x}]

\footnotesize
\textbf{Agents:} ``Yes, based on the architectural style and design of the building in the image, it appears to be Ringkirche...'' \\
\vspace{0.1cm}
\textcolor{green!60!black}{\ding{51} Accurate Visual Verification}
\end{tcolorbox}
\vspace{0.2cm}
\textbf{Final Consensus:} \textcolor{green!60!black}{\ding{51} Correct (Yes)}
\end{minipage}
\end{tcolorbox}
\caption{\textbf{Mitigating Visual Hallucinations.} Qualitative comparison on the MME dataset. While the standard baseline enters a compounding hallucination loop (detailing non-existent flowers and bumblebees), DREAM's DRA module detects high confidence at the $2.0\times$ resolution and routes the debate optimally, yielding the correct answer with high precision.}
\label{fig:qual_mme}
\end{figure*}

\end{document}